\documentclass[runningheads]{llncs}

\usepackage[mobile]{eccv}

\usepackage{eccvabbrv}

\usepackage{graphicx}
\usepackage{booktabs}

\usepackage{algorithm}
\usepackage{algpseudocode}
\usepackage{wrapfig}

\usepackage[accsupp]{axessibility}  % Improves PDF readability for those with disabilities.

\newcommand{\toolname}{\texttt{ArtSavant}}

\usepackage[pagebackref,breaklinks,colorlinks,citecolor=eccvblue]{hyperref}
\usepackage{orcidlink}

\begin{document}

\title{Rethinking Artistic Copyright Infringements in the Era of Text-to-Image Generative Models} 

\titlerunning{Rethinking copyright}

\author{Mazda Moayeri\inst{1} \and
Samyadeep Basu\inst{1}\and Sriram Balasubramanian\inst{1}\and Priyatham Kattakinda\inst{1}\and Atoosa Chengini\inst{1}\and Robert Brauneis\inst{2}\and  
Soheil Feizi\inst{1}}

\authorrunning{M.~Moayeri et al.}

\institute{University of Maryland, Computer Science Department \and
George Washington University, Law Department\\
Correspondence to \email{mmoayeri, sfeizi @umd.edu}}

\maketitle

\begin{abstract}
% %% Step 1: High-level version of the copyright problem and how's it solved currently 
Recent text-to-image generative models such as Stable Diffusion are extremely adept at mimicking and generating copyrighted content, raising concerns amongst artists that their unique styles may be improperly copied.
%While there exists a plethora of recent works on copyright infringements, they primarily focus on either (i) designing methods to make copyrighted training data ``unlearnable''  or (ii) understanding instance-wise copying from generative models. 
Understanding how generative models copy ``artistic style'' is more complex than duplicating a single image, as style is comprised by a set of elements (or \emph{signature}) that frequently co-occurs across a body of work, where each individual work may vary significantly. In our paper, we first reformulate the problem of ``artistic copyright infringement'' to a classification problem over image sets, instead of probing image-wise similarities. We then introduce~\toolname, a practical (i.e., efficient and easy to understand) tool to (i) determine the unique style of an artist by comparing it to a reference dataset of works from $372$ artists curated from WikiArt, and (ii) recognize if the identified style reappears in generated images. We leverage two complementary methods to perform artistic style classification over image sets, including 
TagMatch, which is a novel inherently interpretable and attributable method, making it more suitable for broader use by non-technical stake holders (artists, lawyers, judges, etc). Leveraging~\toolname, we then perform a large-scale empirical study to provide quantitative insight on the prevalence of artistic style copying across 3 popular text-to-image generative models. Namely, amongst a dataset of prolific artists (including many famous ones), {\it only} 20$\%$ of them appear to have their styles be at a risk of copying via simple prompting of today's popular text-to-image generative models. 

%% Step 2: What we do (Linking copyright law ===> Tool)
% In our paper, we draw insights from the law literature, to develop \toolname~-- a practical tool to detect and attribute copyrighted artistic content in generated images through the presence of {\it unique artistic signatures}. 
% %% Step 3: Describe technical aspects of the pipeline
% In particular, \toolname~leverages a curated art dataset from WikiArt, a large bank of artistic concepts from a strong language model (e.g., Vicuna-7B)  along with the capabilities of vision-language models (e.g., CLIP) to design a reliable and interpretable pipeline for artistic copyright detection and attribution.
% %% Step 4: Discuss results and takeways
% Leveraging \toolname, we analyse various copying mechanisms across different text-to-image generative models, highlighting that although generated artistic works are not fully copied at the instance level, they are often copied at an abstract level in the form of the presence of certain {\it artistic signatures}. 
\end{abstract}

\section{Introduction}
\label{sec:intro}
% %% Step 1: High-level version of the copyright problem and how's it solved currently 
In the recent years diffusion-based text-to-image generative models such as Stable Diffusion, Imagen, Mid-Journey, and DeepFloyd \cite{rombach2021highresolution, saharia2022photorealistic, deepfloyd, podell2024sdxl} have captured widespread attention due to their impressive image generation capabilities. Notably, these models demonstrate exceptional performance with very low FID scores on various conditional image generation benchmarks,  showcasing their advanced capabilities.  These models are pre-trained on a large data corpus such as LAION \cite{schuhmann2022laionb} containing up to 5B image-text pairs, which mirror a vast range of internet content, including potentially copyrighted material. This raises an important question - \textit{to what extent do image generative models learn from these copyrighted images?}  While previous studies \cite{carlini2023extracting, somepalli2022diffusion, NEURIPS2023_9521b6e7} have shown that direct copying in diffusion models on the level of individual images is generally rare and mostly occurs due to duplications in the training data, the degree to which image generative models replicate art \textit{styles} as opposed to  art works remains unclear. This issue is increasingly critical as artists express concerns about generative models mimicking their unique styles, potentially saturating the market with imitations and undermining the value of original art. Furthermore, there are no laws currently to identify and protect an artist's style - mainly due to challenges in definition and a previous lack of necessity.

\begin{figure}[t]
    \centering
    \includegraphics[width=\linewidth]{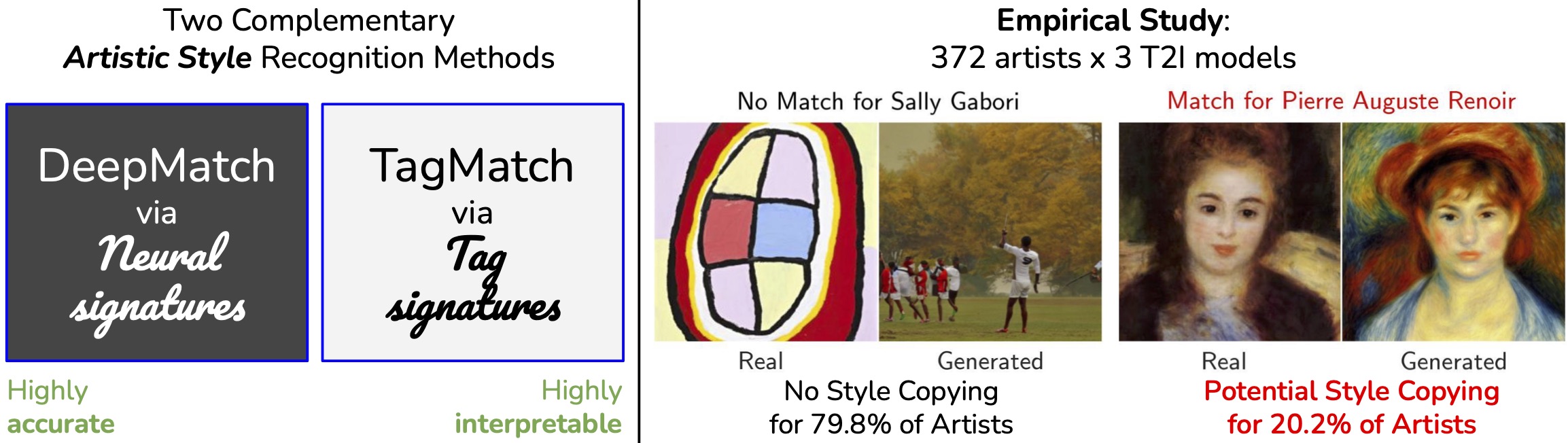}
    \caption{We define artistic style as a set of elements (or signature) that appear frequently over a body of work, and reduce the problem of style copy detection to classification of \emph{sets} of images to artists. ({\bf left}) We propose two ways to recognize artistic styles over a \emph{set} of images, including a novel inherently interpretable and attributable tag-based method. ({\bf right}) In an empirical study of $372$ prolific artists, we find generative models potentially copy artistic styles for $20.2\%$ of these artists.}
    \label{fig:teaser}
\end{figure}

\begin{figure}[t]
    \centering
    \includegraphics[width=\linewidth]{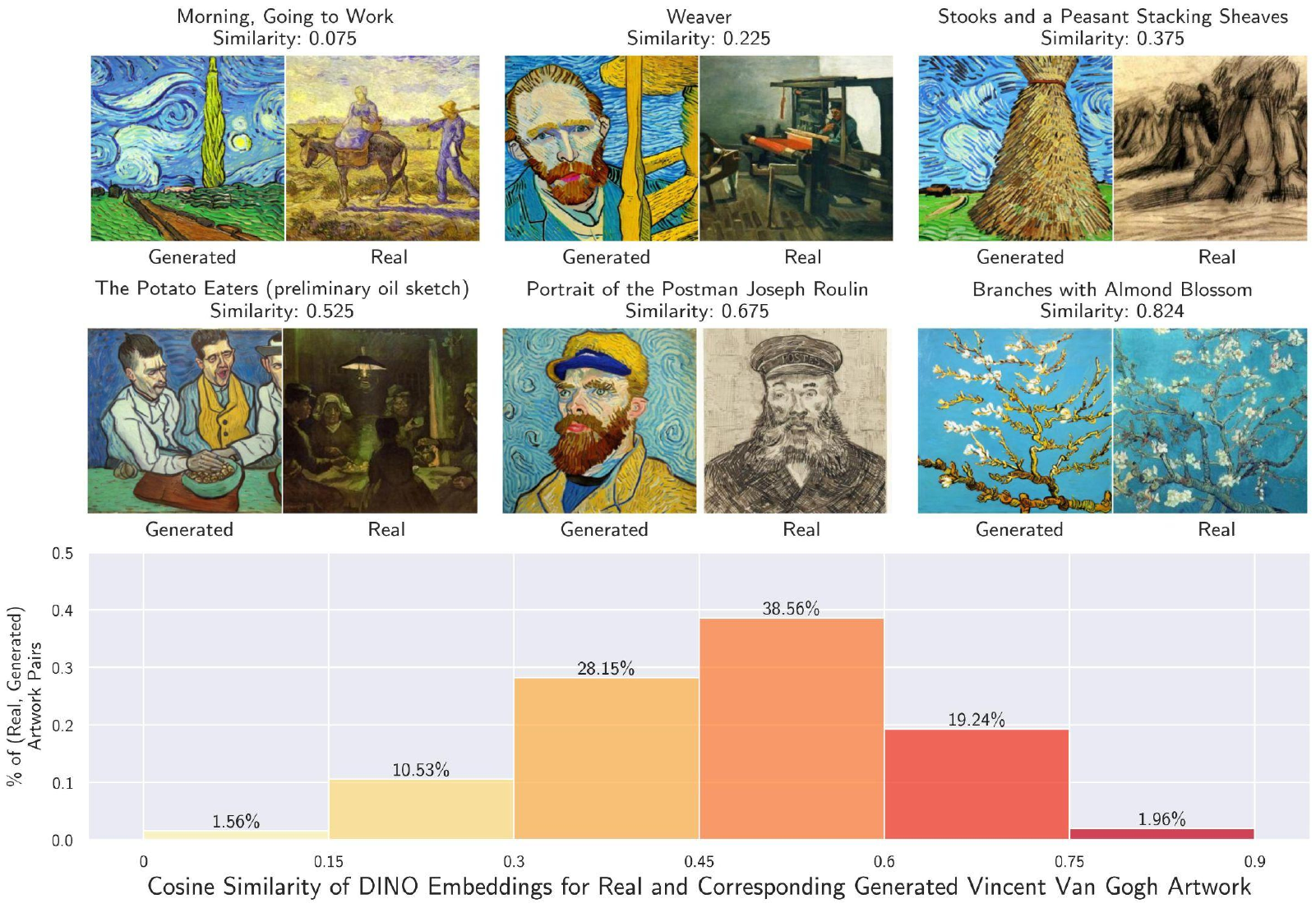}
    \caption{Example generations from Stable Diffusion 2 when prompted to produce specific paintings by Vincent Van Gogh, along with the histogram of similarities between the generated image and corresponding real image. Even for a famous artist like Vincent Van Gogh, generative models rarely produce near-exact duplicates. However, Van Gogh's \emph{style} appears consistently, even when similarity is low.}
    \label{fig:motivation}
\end{figure}

\looseness=-1
Artistic styles are complex, broadly defined over a set of artworks created over their lifetime, making it challenging to determine a style by inspecting individual works of art (a la previous image-wise copy studies). We frame artistic style as characterized by a set of elements that co-occur frequently across works by that artist. For e.g., Vincent Van Gogh had a characteristic art style associated with Post-Impressionism - comprising expressive wavy lines, bright unblended coloring and his signature choppy textured brushwork. In Figure~\ref{fig:motivation}, we illustrate that while these models seldom reproduce Van Gogh's artworks exactly, they frequently capture and replicate elements of his distinctive style. 

To empirically study style copying in generative models and to build a corpus of artistic styles, we first collect an art dataset consisting of artworks from WikiArt from 372 artists, along with the artist labels.  We then proceed to develop \toolname~- a practical tool which can effectively detect and attribute an image to its original artist. The design of this tool is strongly motivated by (i) the notions of `holistic' and `analytic' comparisons from the copyright legal literature \cite{goldsteincopyright, tufenkian} and (ii) shedding insight on the question: {\it Is there a unique set of elements co-occurring across a given artist's works, and if so, can we extract this `signature'?}

For a style to be considered unique, it must be distinguishable from the styles of other artists. However, describing an art style is challenging and making a case for the distinctiveness between two styles is even more so, particularly as artists frequently draw inspiration from each other. An alternative approach to proving the uniqueness of artistic styles involves demonstrating that from a collection of artworks, one can identify the artist(s) who created them. Hence, if art can be accurately attributed to its creators, this would suggest the existence of unique styles that differentiate one artist's works from another's. Therefore, the task of showing the existence and distinctiveness of artistic styles can be reduced to a \textbf{classification problem} (arguably simpler than computing image-similarity).

Our tool's first component -- \textbf{DeepMatch} -- is a neural network which classifies an artwork to its corresponding artist. DeepMatch implicitly maps each artist to a vector (via the classification head) during training, which can be interpreted as a \textit{neural signature} representing an artist. Aggregating its predictions over a set of artworks via majority voting, we find that DeepMatch achieves $89.3\%$ test accuracy. This success indicates that \textit{unique artistic styles do indeed exist for a large fraction of artists}. Since deep features are not very interpretable, DeepMatch is not suited for articulating the elements that comprise each artistic style. For \toolname~to be useful and trustworthy to artists, lawyers, judges or juries, its output must be interpretable by design. Thus, there is a need to complement DeepMatch with a more transparent system which scaffolds a black-box AI component with interpretable intermediate outputs and combines these outputs to arrive at a final prediction. To fill this need, we introduce \textbf{TagMatch}, an inherently interpretable and attributable tag-based classifier. 

TagMatch first tags individual artworks using a tagging method via a novel \textit{zero-shot, selective, multilabel classification} using CLIP \cite{radford2021learning}. These tags are drawn from a concept vocabulary spanning diverse aspects of artistic style, created using LLMs such as Vicuna-13B \cite{zheng2023judging}, and validated using an MTurk human study. They describe a wide variety of artistic attributes like composition, coloring, shapes, and medium. While individual tags are common across artists and thus cannot define unique styles, we propose an efficient algorithm to search over the vast space of combinations over tags to surface \emph{tag signatures}. Namely, we compose common `atomic tags', and find that tag compositions become less frequent as the number of atomic tags within them increases. If a tag combination is unique to a specific artist, it can then be interpreted as a \textit{tag signature} representing a unique artist, and even be used to a set of works to a known artistic style, defined over a portfolio of works for a reference artist.
% For example, we find that {\color{blue} Roerich's artwork consistently contains ``\textit{geometric shapes, simple colors, and geographical symbolism}''. } 
We find such signatures for \emph{all} artists in our dataset, and tag signatures are reliable enought to detect the style of the artists in our dataset (on a held out set) with $61.6\%$ top-1 and $82.5\%$ top-5 accuracy. TagMatch is also highly efficient -- once the CLIP embeddings are computed and cached, it takes just around a minute to search and find tag combinations, despite the search space being combinatorial. But most importantly, with each style detected, TagMatch also articulates the stylistic elements that were uniquely present in the test set of images and the matched reference set. Moreover, TagMatch is attributable, as one can inspect the subset of images from both sets that contain the matched tag signature.

Utilizing both these components, we proceed to use \toolname~to classify the images generated by modern text-to-image models when prompted with an artist's name to the original artists. Somewhat surprisingly, even for the prolific artists we investigate, we find high risk of style copying for only about $20\%$ of artists. We find that while style copying is not highly prevalent for the artists in our dataset, it still occurs very often and may become more common as models become better. 
%For instance, we find that {\it only} 20$\%$ of the artists in our dataset have some level copying in the outputs of generative models. 

In summary, we make the following contributions in our paper:
\begin{itemize}
    \item We reformulate the copyright infringement of artistic styles through the lens of classification over image sets, rather than a single image. 
    \item We introduce~\toolname, a tool consisting of a reference dataset of artworks from $372$ prolific artists, and two complementary methods (including a novel, highly interpretable and attributable one) which effectively can detect unique artistic styles. 
    \item Leveraging \toolname, we perform a large-scale empirical study to understand style copying across 3 popular text-to-image generative models, highlighting that generated images (using simple prompting) from {\it only} 20$\%$ of the artists in our dataset appear to be at high risk of style copying.  
\end{itemize}

\section{Related Works}

As image generative models have rapidly improved in scale and sophistication, the possibility of them mimicking artists' personal styles has been an important topic of discussion in the literature \cite{ren2024copyright}.  
Many previous works describe ways to either detect potential direct image copying in generated images, or to foil any future copying attempts by imperceptibly altering the artists' works to prevent effective training by the generative models. These include techniques like adding imperceptible watermarks to copyrighted artworks \cite{wang2024diagnosis, cui2023diffusionshield, cui2024ftshield}, and crafting ``un-learnable'' examples on which models struggle to learn the style-relevant information  \cite{shan2023glaze, xue2024effective, zhao2023protective}. These methods are typically computationally expensive and incur a loss in image quality, which may render these techniques impractical for many artists. Also, they do not protect artworks which have been previously uploaded to the internet without any safeguards. Others have suggested methods to mitigate this issue from the model owner's perspective - to either de-duplicate the dataset before training \cite{carlini2023extracting, somepalli2022diffusion, NEURIPS2023_9521b6e7}, or to remove concepts from the model after training (``unlearning'') \cite{kumari2023ablating, gandikota2023unified, basu2023localizing}. These are also technically challenging, and require the model owner to invest significant resources which may again inhibit their practicality. Methods like \cite{carlini2023extracting, somepalli2022diffusion, NEURIPS2023_9521b6e7} are also more focused on analyzing direct image copying from the training data, and thus may not be applicable to preventing style copying.

None of these works tackle the problem of \textit{detecting} potentially copied art \textit{styles} in generated art, especially in a manner which may be relevant to legal standards of copyright infringement. According to current US legal standards \cite{CRS_Reports_2023}, an artwork has to meet the ``substantial similarity'' test for it to be infringing on copyright. This similarity has to be established on \textit{analytic} and \textit{holistic} terms \cite{tufenkian, goldsteincopyright}. {Analytic} here refers to explaining an artwork by breaking it down into its constituents using a concrete and objective technical vocabulary, while {holistic} refers to the overall ``look and feel'' of the artwork. Thus, it is highly desirable for any automated copy-detection method to reflect this dichotomy in its working. We thus design our tool based on these two notions.

\section{Motivation}
Recent works have investigated copying on an image-wise level, showing that state-of-the-art diffusion models can generate exact replicas of a small fraction of training set images ~\cite{somepalli2022diffusion, NEURIPS2023_9521b6e7, carlini2023extracting}. Typically, these works involve representing images in a deep embedding space via models like SSCD~\cite{pizzi2022selfsupervised} or DINO~\cite{DBLP:journals/corr/abs-2104-14294}, and computing image-to-image similarities across generated and real images. These results, as well as anecdotal instances, have raised concerns amongst artists, since generative models may pose a risk of saturating the market with replications, thus jeopardizing the artists' livelihoods, as well as cheapening creative work they likely feel personal ownership and attachment to. Inspired by these valid concerns and existing results on image replication, we first explore if generative models can recreate famous artworks, e.g., by Vincent Van Gogh. Specifically, we generate images by prompting ``{\it\{artwork title\}} by Vincent Van Gogh'' for $1500$ Van Gogh works, and compute the DINO similarity between correponding real and generated pairs. Figure \ref{fig:motivation} visualizes the distribution of similarities, as well as examples at each similarity level. We find that the vast majority of similarities are lower than $0.75$, which corresponds to (generated, real) pairs that are far from duplicates. However, even when the generated image differs significantly from the source real image, certain stylistic elements associated with Van Gogh seem to appear consistently in the generated works. 
%In fact, some of the low similarity pairs occur because the generated image utilizes Van Gogh's most famous style, while the real painting features lesser known sub-styles or mediums (e.g. a sketch). 
Thus, while instance-wise copying of artwork appears rare for even the very famous Van Gogh, we argue that style copying requires going beyond image-to-image comparisons, as artists may still have their personal styles, developed over a long time and at significant personal cost, infringed upon in ways that searching for exact replicas would miss. Namely, to investigate style copying, one must first identify one (or more) styles used by an artist, which we define as a set of elements, or \emph{signature}, that frequently occurs across a multiple works.  

Currently, copyright law does not protect an artist's style; however, this may be due to the prior absence of a pressing need for such protection from humans, combined with the difficulty of defining a person's style. Given the rapid advance of generative models, we attempt to tackle the problem of defining and identifying artistic styles. Importantly, so that our results are useful to a broad audience, we prioritize transparency and efficiency in our work. That is, we design a tool that is fast enough for an end-user (e.g., artist or lawyer) to run, and interpretable enough so that the user can understand and convey the results to another party (e.g., judge or jury). In line with the existing notions of `analytic' and `holistic' components of an artwork, we extend these notions to an artist's style, which may be developed over a \textit{set} of artworks comprising an \textit{art portfolio}. Here, the holistic component would refer to the overall `look and feel' of the artist's style, while the analytic component would consist of a breakdown of an art style into its constituent components. We make this concrete by designing a style copy detection tool which corresponds closely with the mentioned concepts and operates on a \textit{set} of images rather than on a single image. 

\vspace{-0.3cm}
\section{Towards Practical Artistic Style Copy Detection}
To argue an artist's style is copied, one must first demonstrate the existence of a unique style for the artist. An \emph{analytic} approach is to articulate the frequently co-occurring elements that comprise the artist's style. Alternatively, a \emph{holistic} argument is to show that the artist's work can consistently be distinguished from that of other artists, than there must exist something unique that is present across the artist's portfolio. In the latter case, we have reduced style copy detection to a classification problem over \emph{sets} of images (i.e. artist portfolios), something neural networks are well suited to do. We now propose DeepMatch and TagMatch, two complementary methods (w.r.t. accuracy and interpretability) that detect artistic styles in holistic and analytic manners, respectively. 

% For a style to be considered unique, it must be distinguishable from the styles of other artists. However, describing an art style is challenging and making a case for the distinctiveness between two styles is even more so, particularly as artists frequently draw inspiration from each other. An alternative approach to proving the uniqueness of artistic styles involves demonstrating that from a collection of artworks, one can identify the artist(s) who created them. Hence, if art can be accurately attributed to its creators, it suggests the existence of unique styles that differentiate one artist's works from another's.

% Therefore, the task of showing the existence and distinctiveness of artistic styles can be reduced to a classification problem. Specifically, this involves a task where, given a collection of test images, one must determine the correct style from among several known styles, each represented by its own set of images.

% Thus, we introduce (i) a curated dataset comprising of artworks from 386 artists which contains a broad variety of known styles, and (ii) two complementary methodologies for executing this style classification task which are aligned with the `holistic' and `analytic' modes of thinking from the copyright legal community.

\subsection{WikiArt Dataset}
To distinguish on artist's style from that of others, we need a corpus of artistic styles (consisting of portfolios from many artists) to compare against. To this end, we curate a dataset $\mathcal{D}$ consisting of artworks from WikiArt~\footnote{https://www.wikiart.org/} to serve as (i) a reference set of artistic styles, (ii) a validation set of real art to show our method's can recognize an artist's style when shown a held-out set of their works, and (iii) a test-bed to explore if text--to-image models replicate the styles of the artists in our dataset in their generated images. Previous work \cite{DBLP:journals/corr/TanCAT17} uses images from WikiArt to for a different purpose (GAN training). Since the content of WikiArt has been updated since then, we re-scrape WikiArt, perform a filtering step and subsequently curate a repository of $\sim$91k painting images encompassing 372 unique artists (denoted by the set $\mathcal{A}$). The filtering step ensures that each artist $a \in \mathcal{A}$ has at least 100 images. 
We also denote the set of images (or portfolio) for each artist $a \in \mathcal{A}$ as $\mathcal{D}_{a}$. Each of the images in our dataset is annotated with its corresponding genre (e.g., {\it landscape}) and style (e.g., {\it Impressionism}) which provides useful information about the paintings beyond their given titles. We provide an easy-to-execute script with all the necessary filtering step, so to generate newer versions of WikiArt if desired (see Appendix). We now detail our two complementary methods that compare a test set of images to our reference corpus so to detect if any of the reference styles reappear. 

\vspace{-0.4cm}
\subsection{DeepMatch: Black-Box Detector}
\label{subsec: style_clip}
\begin{wrapfigure}{l}{0.475\linewidth}
    \vspace{-0.9cm}
    \centering
    \includegraphics[width=\linewidth]{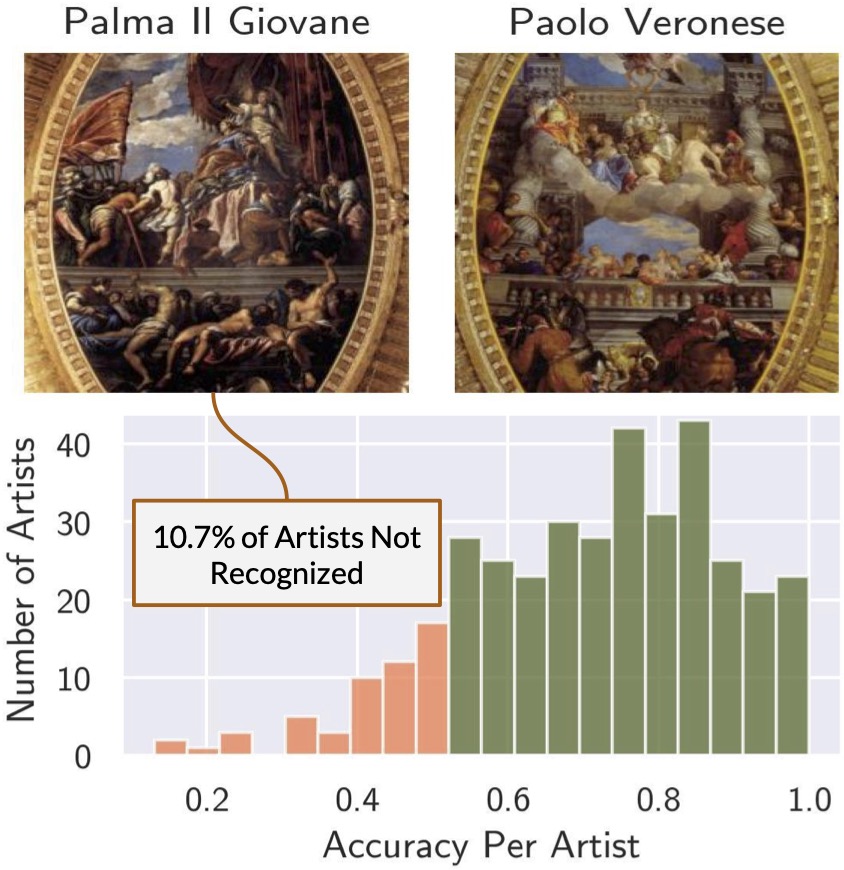}
    \caption{DeepMatch on held-out real art: $89.3\%$ of artists can be recognized. The remaining $10.7\%$ of artists have very similar styles to other artists: e.g., Palma Il Giovane's work differs marginally from other Italian renaissance painters.}
    \label{fig:unrecog_artists}
    \vspace{-0.8cm}
\end{wrapfigure} DeepMatch consists of a light-weight artist classifier (on images) and a majority voting aggregation scheme to obatin one prediction for a \emph{set} of images. Majority voting requires that at least half the images in a test set $\hat{D_a}$ are predicted to $a$ for DeepMatch to predict $a$, allowing for abstention in case no specific style is recognized with sufficient confidence. For our classifier, we train a two layer MLP on top of embeddings from a frozen CLIP ViT-B\textbackslash16 vision encoder \cite{radford2021learning} to classify artwork to their respective artist, using a train split containing $80\%$ of our dataset. We employ weighted sampling to account for class imbalance. Because we utilize frozen embeddings, training is very fast, taking only a few minutes. Thus, a new artist could easily retrain a detector to include their works (and thus encode their artistic style).
 % artwork to their respective artists. We train the MLP (with a 1024 dimensional hidden layer) using Adam with a learning rate of 0.001 for 35 epochs on the train split containing roughly $73k$ images ($80\%$ of the dataset).
% In our experiments with various backbones, we find that CLIP ViT-??? produces the highest classification accuracy over artists - with a top-1 accuracy of 73\% on the test set. This indicates the existence of an artistic style for at least a significant portion of the WikiArt Dataset.

\textbf{Validation of the Detector. } We apply DeepMatch on the held-out test split of our dataset and observe that the image-wise classifier attains $72.8\%$ accuracy per image over $372$ artists. When aggregating image-wise predictions via majority vote, $89.3\%$ of artists are matched, validating our method, and offering strong evidence towards the existence of unique artistic styles. Specifically, neural classifiers capture unique and frequently co-occurring characteristics of the artists in their embedding space, which can be thought of as \emph{neural signatures}. Figure \ref{fig:unrecog_artists} shows the distributions of image-wise accuracies per artist, shading correctly matched images (green). We also present an image from one of the few artists who's style is not matched by DeepMatch, along with an image from a similar artist. Notice that the style of two artists can be extremely similar, making the existence of unique artistic styles for the vast majority of artists considered (by way of neural signatures) a non-trivial observation.

% In our experiments with various backbones, we find that CLIP ViT-B/16 produces the highest classification accuracy over artists - with a top-1 accuracy of 73\% on the test set. This indicates the existence of an artistic style for at least a significant portion of the WikiArt Dataset. In particular,~\Cref{fig:unrecog_artists} shows that a significant number of artists can be identified using a set of their works. This shows that neural classifiers indeed capture some characteristics of the artists in their embedding space in the form of neural signatures. However, these neural signature are not interpretable. To address this issue, in the next section, we discuss an efficient method to obtain interpretable signatures corresponding to artists. 
\subsection{Interpretable Artistic Signatures}
\label{subsec: artist_signatures}
Now we provide an interpretable alternative to matching via neural signatures. We draw inspiration from the interpretable failure mode extraction of \cite{rezaei2023prime}. Namely, we first tag images with atomic tags drawn from a vocabulary of stylistic elements. Then, we \emph{compose} tags efficiently to go from atomic tags that are common across artists to longer tag compositions that are unique to each artist (i.e. \emph{tag signatures}). Lastly, we detect artist styles in an attributable way via match tag signatures in a test portfolio to those for artists in our reference corpus. We detail each step of TagMatch, as well as validation results, below. 

\begin{figure}[h]
    \centering
    \begin{minipage}{0.49\linewidth}
    \includegraphics[width=\textwidth]{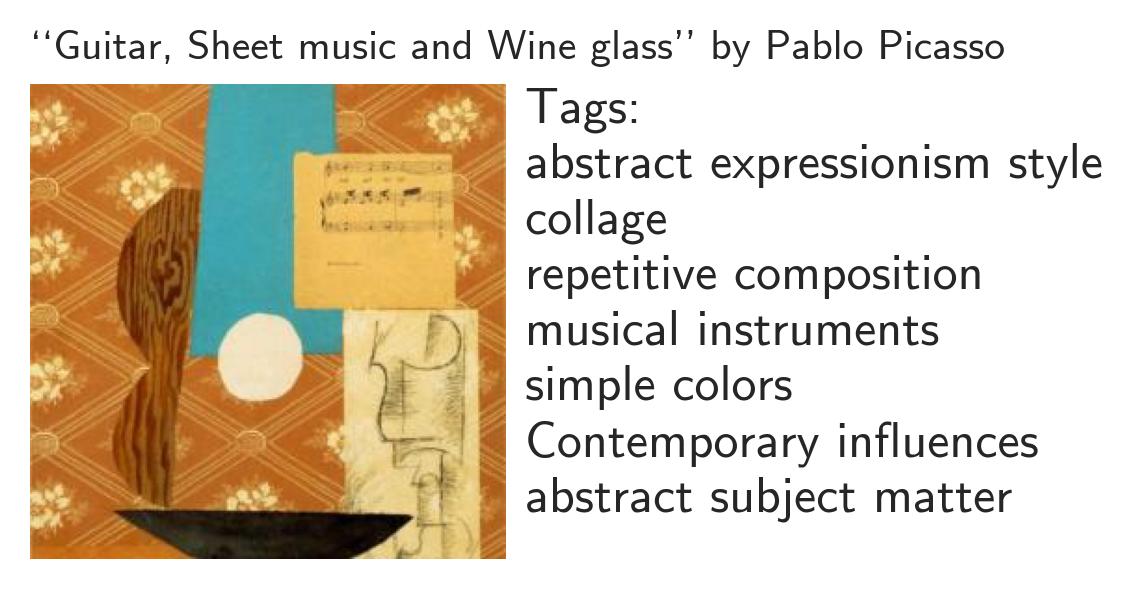}
    \end{minipage}
    \begin{minipage}{0.49\linewidth}
    \includegraphics[width=\textwidth]{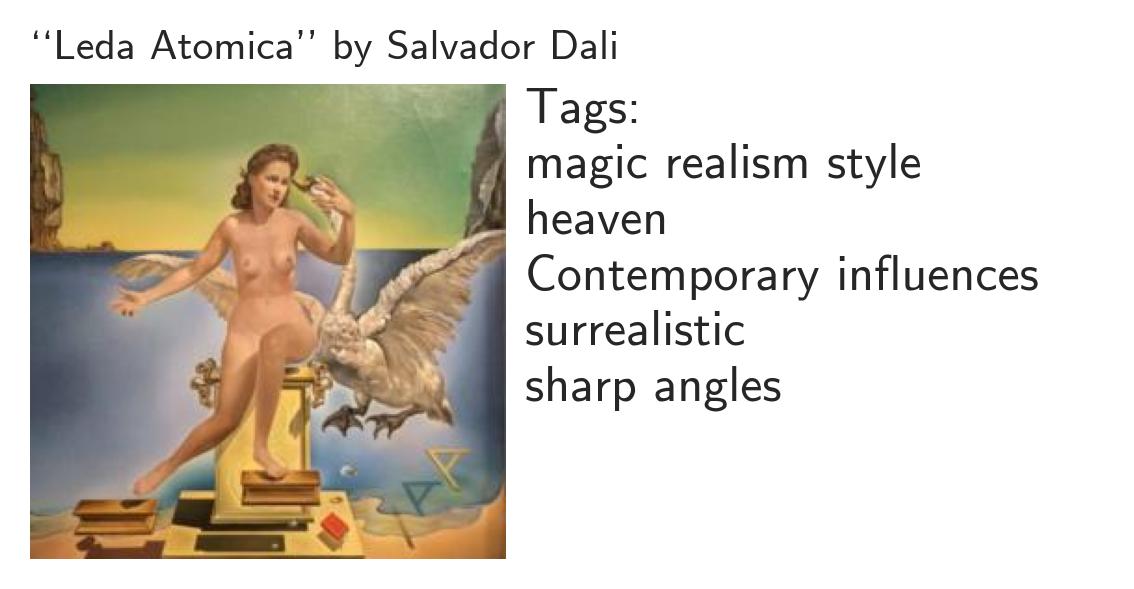}
    \end{minipage} \\
    \begin{minipage}{0.49\linewidth}
    \includegraphics[width=\textwidth]{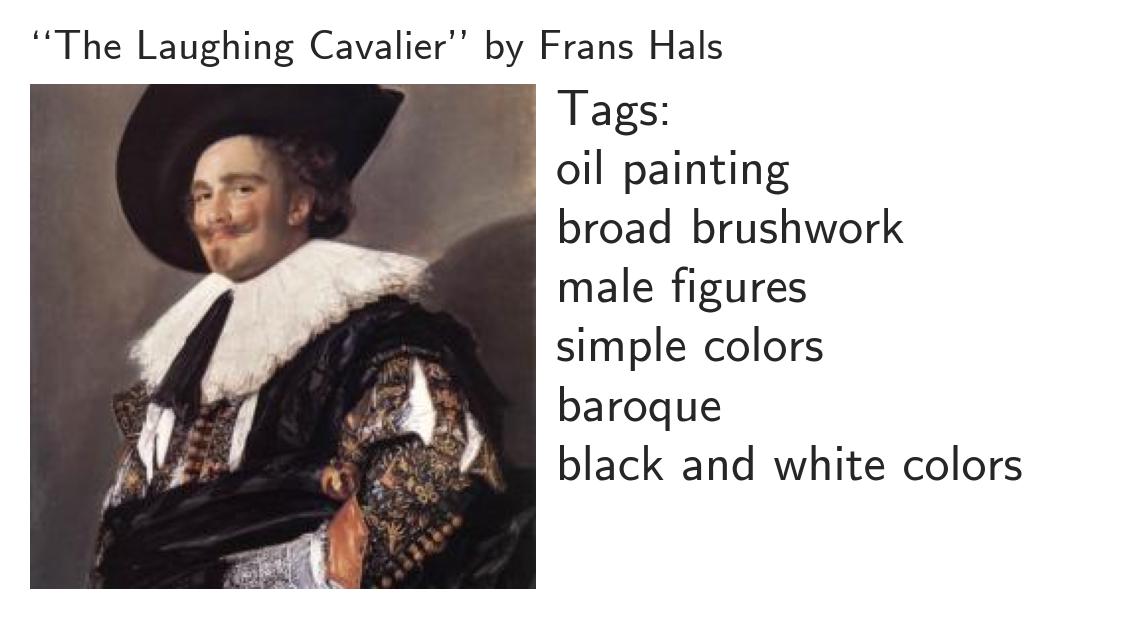}
    \end{minipage}
    \begin{minipage}{0.49\linewidth}
    \includegraphics[width=\textwidth]{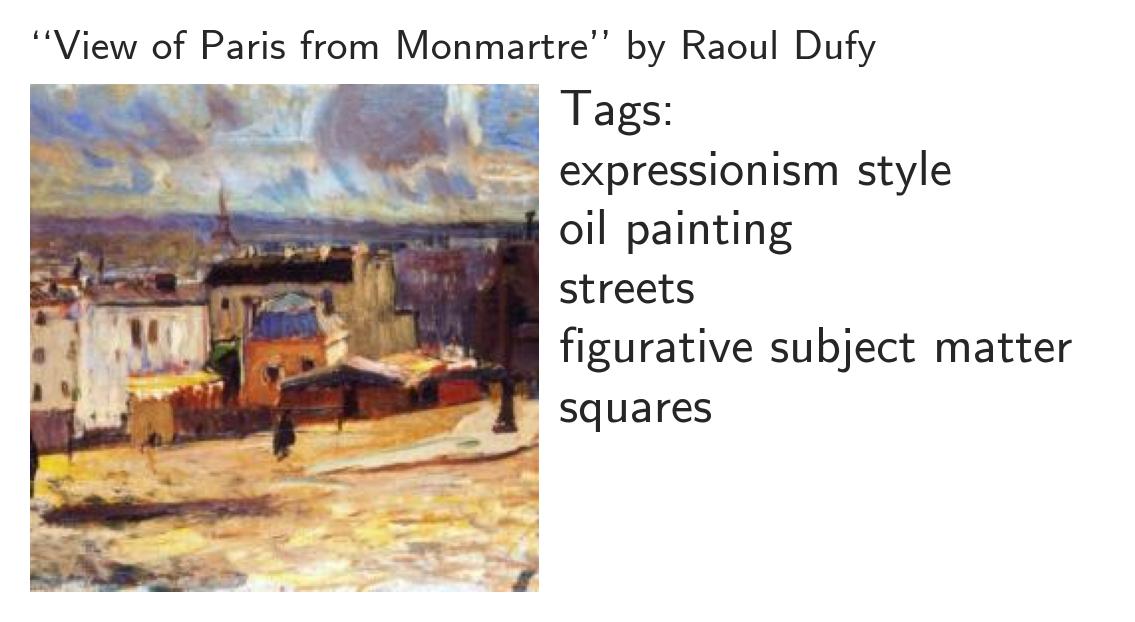}
    \end{minipage}
    \caption{Example atomic tags assigned via our proposed CLIP-based zero-shot method. We perform selective multilabel classification along various aspects of art (e.g. medium, colors, shapes, etc), so that atomic tags span diverse categories. Details in section \ref{subsec:tagging}. }
    \label{fig:eg_tags}
\end{figure}

\textbf{Zero-shot Art Tagging}
\label{subsec:tagging}
We utilize the zero-shot open-vocabulary recognition abilities of CLIP to tag images with descriptors of stylistic elements. First, we construct a concept vocabulary $\mathcal{V}$ with help from LLMs. Namely, we prompt Vicuna-13b and ChatGPT to generate a dictionary of concepts along various aspects of art. We manually consolidate and amend the concept dictionary, resulting in a vocabulary of $260$ concepts over $16$ aspects (see Appendix \ref{app-sec:atomic_tags}). %We call these concepts \emph{atomic tags}. 

To assign concepts to images, we a design a novel scheme that consists of selective multilabel classification per-aspect. Namely, for an image, we compute CLIP similarities to all concepts, and normalize similarities \emph{within each aspect}. Then, we only assign a concept its normalized similarity (i.e. z-score) exceeds a threshold of $1.75$. This means that a concept is only assigned for an aspect if the image is substantially more similar to this concept than other concepts describing the same aspect. Classifying per-aspect allows for a diversity of descriptors to emerge, as global thresholding results in a biased tag description, as concepts for certain aspects (e.g. subject matter) consistently have higher CLIP similarity than those for more nuanced aspects (e.g. brushwork). We call the assigned concepts \emph{atomic tags}; figure \ref{fig:eg_tags} shows atomic tags assigned for a few examples. 

\begin{algorithm}[t]
\caption{Iterative Algorithm to Obtain Tag Composition Per Artist $a \in \mathcal{A}$}\label{alg:cap}
\begin{algorithmic}
\Require $\mathcal{D}_{a}$ (Images for artist $a$), $\mathcal{C}_{a}$ (Common tags for artist $a$) 
\State $\mathcal{S}_{a} = \{ \}$  \Comment Stores the tag compositions with their associated counts
%\While{$x \in \mathcal{D}_{a}$}
\For {$x \in \mathcal{D}_{a}$}
    \State $I(x) = \text{tag}(x) \cap \mathcal{C}_{a}$  \Comment{Compute the intersection with common atomic tags}
    \State $\mathcal{P}(I(x)) = \text{ComputePowerSet($I(x)$)}$  \Comment{Compute power-set of the tags}
    \State UpdateCount($\mathcal{S}_{a}, \mathcal{P}(I(x))$) \Comment Update the count of each tag composition
%\EndWhile
\EndFor 
\State Filter($S_{a}$) \Comment Keep tag compositions which occur above a count threshold of 3
\end{algorithmic} 
\end{algorithm}

\textbf{Validation of Quality of Tags Using Human-Study.} We validate the effectiveness of our tagging via a human-study involving MTurk workers. In particular, given an image of an artwork and an assigned atomic tag $v_{predict}$ from the vocabulary $\mathcal{V}$ -- MTurk workers are asked ``{\it Does the term $v_{predict}$ match (i.e. the concept $v_{predict}$ present) the artwork below?} ''. The workers are then asked to select between $\{ \text{Yes, No, Unsure}\}$. We collect responses for $1000$ images with $3$ annotators each. We find that in only 17$\%$ cases, a majority of workers disagree with the provided tag, suggesting our tagging results in a low false positive rate. We also observe all three annotators agree in only $51\%$ of cases, reflecting that describing artistic style can be subjective. While our tagging is not perfect, it is a deterministic and automatic method of articulating artistic style elements, and that our tagging method will improve as underlying VLMs improve too. See the appendix for more details and discussion on the human study.  
% This result highlights that our tagging approach is effective thereby resulting in a low false positive rate. We provide more details on the human-study template in the Appendix.  

\textbf{Tag Composition for Artists.} Using the atomic tags in the artwork specific vocabulary $\mathcal{V}$, in this section 
we design a simple and easy-to-understand iterative algorithm to obtain a set of \emph{tag signatures} $\mathcal{S}_{a}$ for each artist $a \in \mathcal{A}$. These signatures are a composition of a subset of tags in $\mathcal{V}$.  In particular, our algorithm efficiently searches the space of tag compositions to go from atomic tags to composition of tags which become more unique as the length of the tag composition grows. For e.g., while 40$\%$ of the artists may use simple colors, {\it only} 15$\%$ may use both simple colors and impressionism style. 

To efficiently search the space of tag compositions per artist $a \in \mathcal{A}$, we first assign a set of tags to each of their images $x \in \mathcal{D}_{a}$ via the zero-shot {\it selective multi-label classification} method described above. 
For each image $x$, let $\text{tag}(x)$ denote the set of predicted atomic tags. 
%These sets of initial tags per artist are filtered by keeping only the tags occurring at least $t$ times. 
To get atomic tags \emph{for an artist}, we aggregate all atomic tags over images, and keep only the tags occurring in at least $3$ works. We denote this aggregate set of atomic tags as the ``Common Atomic Tags Per Artist'' and denote it as $\mathcal{C}_{a}$. Then, we iterate through all the images $x \in \mathcal{D}_{a}$ for a given artist $a$, to find the intersection $I(x) = \text{tag}(x) \cap \mathcal{C}_{a}$. We then compute a powerset $\mathcal{P}(I(x))$ of the tags occurring in the intersection $I(x)$ and increment the count of each occurrence of the tag composition from the powerset in $\mathcal{S}_{a}$. Note that the size of $I(x)$ is much smaller than that of $\mathcal{C}_{a}$, and thus, iterating through $\mathcal{P}(I(x))$ for each image $x$ is much, much faster than iterating through $\mathcal{P}(\mathcal{C}_{a})$. Finally, we again filter the tag compositions in $\mathcal{S}_{a}$, only including those that occur in at least $3$ works. We provide the details of this tag composition algorithm in~\Cref{alg:cap}, and discuss other details in the Appendix.

\textbf{Do Unique Signatures Exist for Artists?} Using our tag composition
\begin{wrapfigure}{r}{0.475\linewidth}
    \centering
    \includegraphics[width=\linewidth]{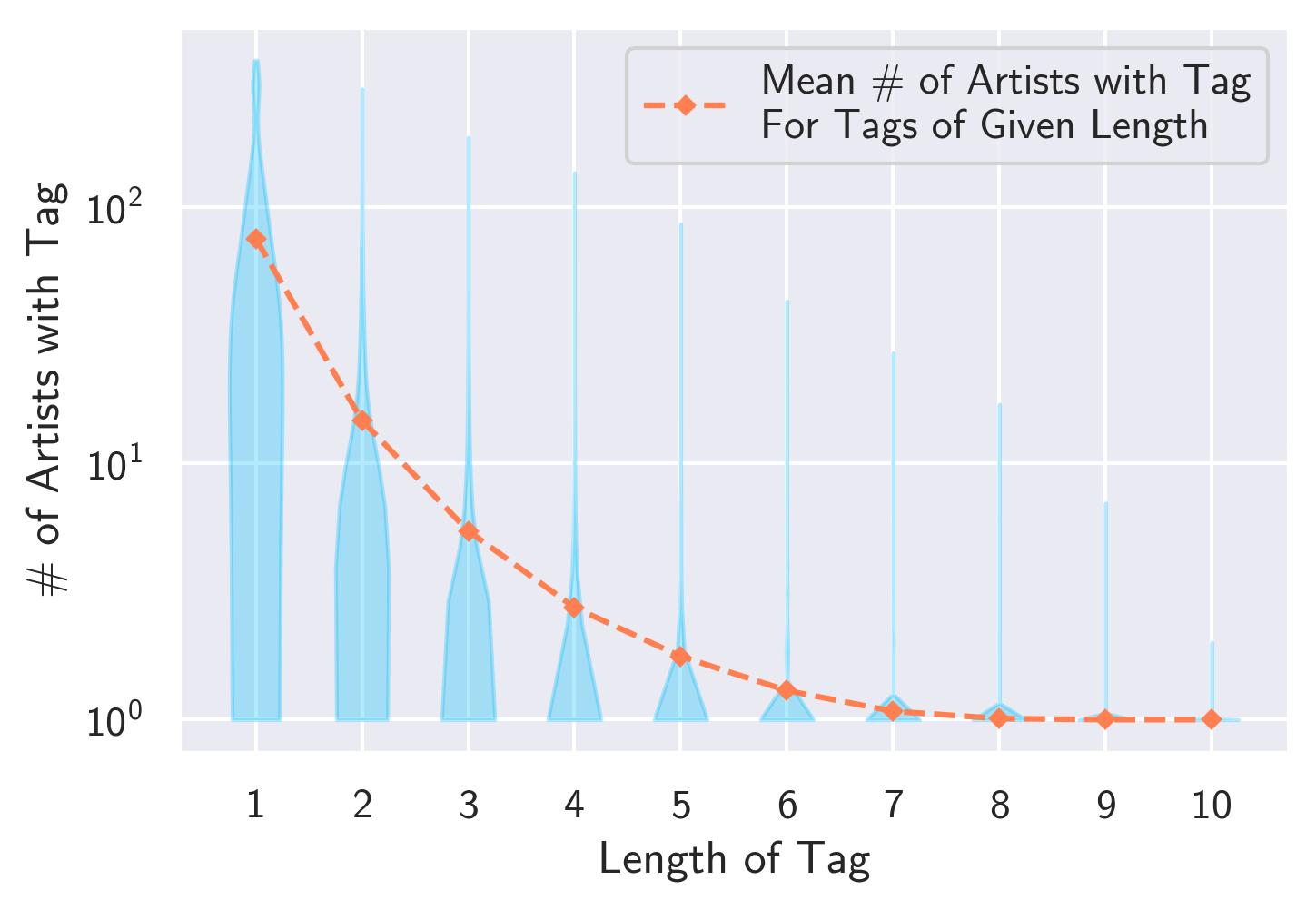}
    \caption{Composing atomic tags results in more unique tags, towards artistic \emph{signatures}. We propose an efficient algorithm to count composed tags; see Algorithm \ref{alg:cap}.}
    \label{fig:tag_composition}
    \vspace{-0.4cm}
\end{wrapfigure}
method on the curated dataset from WikiArt, we find that unique {\it artistic signatures} in the form of an unique tag composition exists per artist. In~\Cref{fig:tag_composition}, we show that our tag composition algorithm is able to select unique tag compositions such that {\it only} a very few artists exhibit such compositions in their paintings as the tag length increases. This shows that artists exhibit {\it unique style} which can effectively be captured by our iterative algorithm. Leveraging these observations, in the next section, we describe TagMatch, which can classify a set of artworks to an artist by uniquely matching such tags (or tag signatures). 
\subsection{TagMatch: Interpretable and Attributable Style Detection}
\label{sec: tagmatch}
In~\Cref{subsec: style_clip}, we outlined a holistic approach to accurately detect artistic styles. While DeepMatch obtains high accuracy (recognizing styles for $89.3\%$ of artists), the neural signatures it relies upon lack interpretability. For a copyright detection tool to be useful in practice (e.g., to be used as assistive technologies), providing explanations of the classification decisions can tremendously benefit the end-user. To this end, we leverage our efficient tag composition algorithm as defined in~\Cref{subsec: artist_signatures} to develop TagMatch - an interpretable classification and attribution method which can effectively classify a set of artworks to an artist, as well provide reasoning behind the classification and example images from both sets that present the matched tag signature. TagMatch follows the intuition of matching a test portfolio to a reference artist who's portfolio shares the most unique tag signatures. Given a set of $N$ test images $\mathcal{T}=\{x_{i}\}_{i=1}^{N}$, we first obtain a number of tag compositions for them using our iterative algorithm in~\Cref{subsec: artist_signatures}. These tag compositions are then compared with the tag compositions of the artists in the reference corpus in order of uniqueness (i.e. we first consider tag signatures present in the test portfolio that occur for the fewest number of reference artists). We can then rank reference artists by how unique the shared tags are with the test portfolio. Detailed steps of the algorithm is in Appendix (Algo. 2). Altogether, TagMatch is remarkably fast, taking only about a minute, after caching embeddings of all images. 

\textbf{Validation of TagMatch.} We again utilize the test split of our WikiArt Dataset to validate the proposed style detection method. We observe TagMatch to predict the correct artist with top-1 accuracy of $61.6\%$, with top-5 and top-10 accuracies rising to $82.5\%$ and $88.4\%$ respectively. While less accurate than DeepMatch, the \emph{tag signatures} provided by TagMatch allow for analytic arguments to be made regarding style copying, as the exact tag signatures used in matching can be inspected. Moreover, the subset of images in both the test portfolio and matched reference portfolio can be easily retrieved, offering direct attribution of the method; examples can be seen in the next section, where we match generated images to our reference artists. Overall, we hope TagMatch and DeepMatch can serve as automatic and objective tools to navigate the subtle problem of identifying artistic styles, towards understanding when styles are copied and helping artists argue their case (i.e. in a court of law) in such instances. 

% In~\Cref{fig:tag_match}, we show qualitative examples where a set of test images are effectively matched to a reference artist. In particular, we show that each classification decision is associated with the ``Matched Tag'', therefore making our algorithm inherently interpretable. Overall, TagMatch obtain an accuracy of 60$\%$ over different sets of test images from the WikiArt dataset, highlighting its effectiveness. We note that while TagMatch lags behind DeepMatch in terms of accuracy, its inherent interpretable nature makes it more suitable to be useful in practice where trust and reliability of predictions are important, like in a court of law. 

% \section{Experiments}
% {\color{blue} TODO: Subsections here can have the title of the takeaways: (i) Detection Results (Without interpretability) ; (ii) Detection Results (With Interpretability) (iii) Discussion on when to use (i) vs. (ii); (iv) Validation of Tagging with Human-study;}

\begin{figure}
    \centering
    \includegraphics[width=\linewidth]{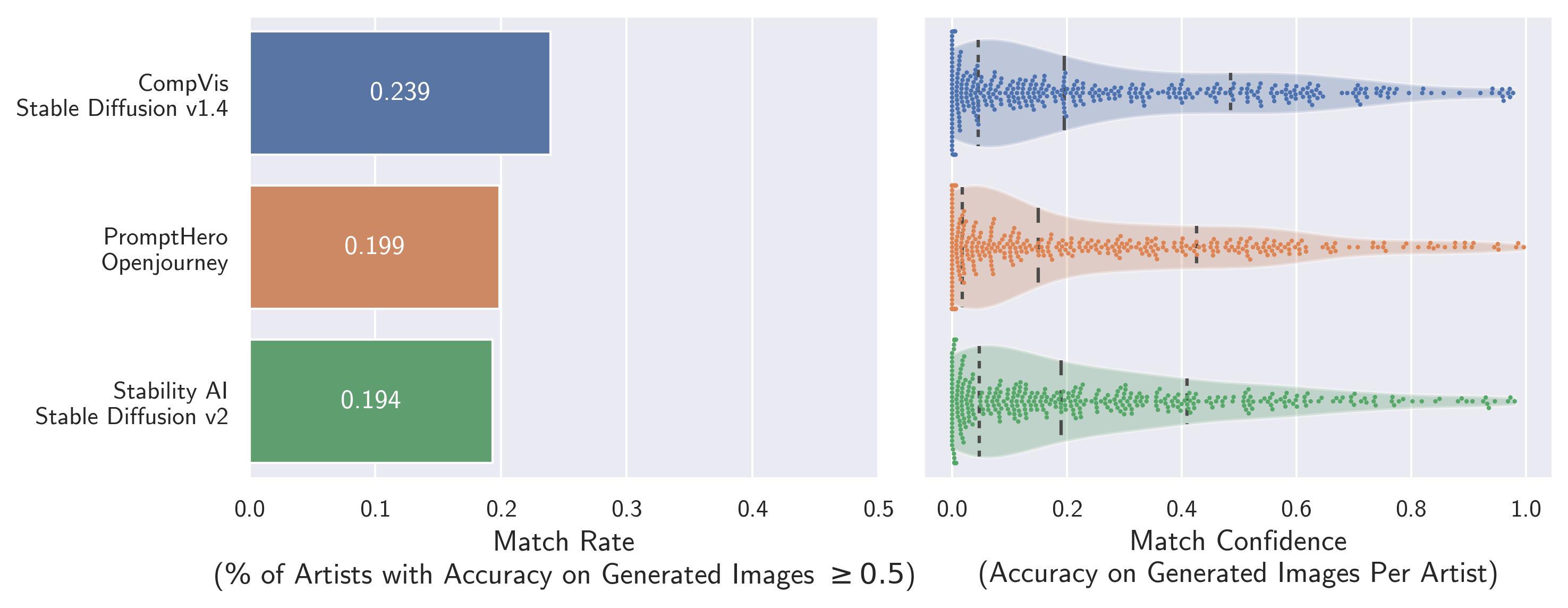}
    \caption{({\bf left)} On generated images, DeepMatch recognizes the intended artist for an average of $20.2\%$ of artists. ({\bf right}) Match confidences range substantially (each point corresponds to the recognition rate for one artist; dotted lines correspond to quartiles), with some artists having no generated images predicted to them, while others have their neural style signatures recognized over $80\%$ of the time ({\bf right}).}
    \label{fig:gen_imgs_artist_recog}
    \vspace{-0.2cm}
\end{figure}
\section{Analysis on Generated Art from Text-to-Image Models}
We now turn to generated images, towards two ends. First, we seek to demonstrate the tools we validated on real art can be similarly effective in recognizing and articulating artistic styles in generated art. Secondly, by conducting a systematic empirical study, we aim to shed quantitative insight into the phenomena of style infringement by generative models. While enough instances of style mimicry have been observed to raise concern \cite{shan2023glaze, ren2024copyright}, the prevalence and nature of such instances remains nebulous. We hope our analysis can provide a more complete picture of the current state of style copying by generative models. 

Specifically, we employ TagMatch and DeepMatch to generated versions of the art in our WikiArt dataset, so to quantify the degree to which generative models reproduce the stylistic signatures of the $372$ artists in our dataset. These artists are somewhat representatitve in the sense that they touch a wide spectrum of broader styles, and they are each somewhat popular and prolific (with respect to having at least $100$ works on WikiArt), making them good candidates to potentially have their styles infringed by generative models.

% Toward this end, we develop language to assist in disambiguating instances where generative models take inspiration from existing art, as any artist does, from cases where unique styles of an artist are replicated with \emph{unprecedented} similarity. While outlawing the former would stifle creativity and hinder artistic liberty, allowing for the latter would hurt artists, as generative models could divert revenue that would otherwise go to the artist's who's unique styles they reproduce. 

% Namely, we demonstrate how TagMatch and DeepMatch can be used to (i) identify and articulate unique styles of an artist that are being infringed, (ii) show that a set of generated images is most similar to the work of an artist, and (iii) the similarity we hope this analysis provides a more empirically-grounded picture of the current state of potential style infringements. 

\textbf{Setup.} We extract the titles from the paintings in our dataset from WikiArt and augment them with the name of the artist. Using these prompts (e.g.``\emph{the starry night} by \emph{Vincent Van Gogh}'' or ``\emph{the water lillies} by \emph{Claude Monet}''), we generate images from 3 text-to-image generative models: (i) Stable-Diffusion-v1.4; (ii) Stable-Diffusion-v2.0; and (iii) OpenJourney from PromptHero. We note that (i) and (ii) are pre-trained on a subset of the LAION dataset~\cite{schuhmann2022laion5b}, while (iii) is pre-trained on LAION and then fine-tuned on Mid-Journey generated images. We also note that (ii) uses a stronger CLIP text-encoder which can help generating images with better fidelity to the text-prompt. These characteristics make these generative models unique to one another, thereby providing a diverse range of artistic interpretations and styles in our image generation experiments.

Thus, for each artist $a_i$ in our dataset, in addition to a set of his or her real artworks $\mathcal{D}_{a_{i}}$, we obtain a corresponding set of generated images $\mathcal{D}_{a_{i}}'$, per generative model. We then compare each set of generated art to the entire corpus of existing art. Namely, we seek to quantify the frequency with which generated art prompted to be in the style of a specific artist is matched to that artist; we call this the \emph{match rate}. Match rate is a percentage over $372$ artists, as each artist is either matched correctly or not (i.e. to the wrong artist or no artist at all). We also consider top-5 and top-10 match rates, where a top-k match refers occurs when artist $a_i$ is amongst the top $k$ predictions for the set $\mathcal{D}_{a_{i}}'$ generated using prompts of the form ``\{title of a work by $a_i$\} by \{$a_i$\}''. 

% That is, given a classifier $f:A \rightarrow \mathbb{R}^d$, where $d=386$ is the number of artists in our dataset, match rate is simply $|{i s.t. 

\subsection{DeepMatch Recognizes Artistic Styles for Half of the Artists}

\looseness=-1
We first employ DeepMatch, the more accurate but less interpretable of our two style recognition systems, to quantify the degree to which the unique styles of artists from our dataset are reproduced by generative models prompted to recreate works from these artists. Recall that DeepMatch predicts an artist from a set of test images by first inferring the artist for each image, and then aggregating predictions across the set. Thus, in addition to a match rate, we can investigate a match \emph{confidence}, which is simply the fraction of images in $\mathcal{D}_{a_{i}}'$ predicted to $a_i$; that is, the accuracy of DeepMatch for a set of generated images $\mathcal{D}_{a_{i}}'$, if we take the label of these images to be artist who's style they attempt to recreate (i.e., $a_i$). Note that match confidence is computed per artist. 

\begin{figure}[t]
    \centering
    \includegraphics[width=\linewidth]{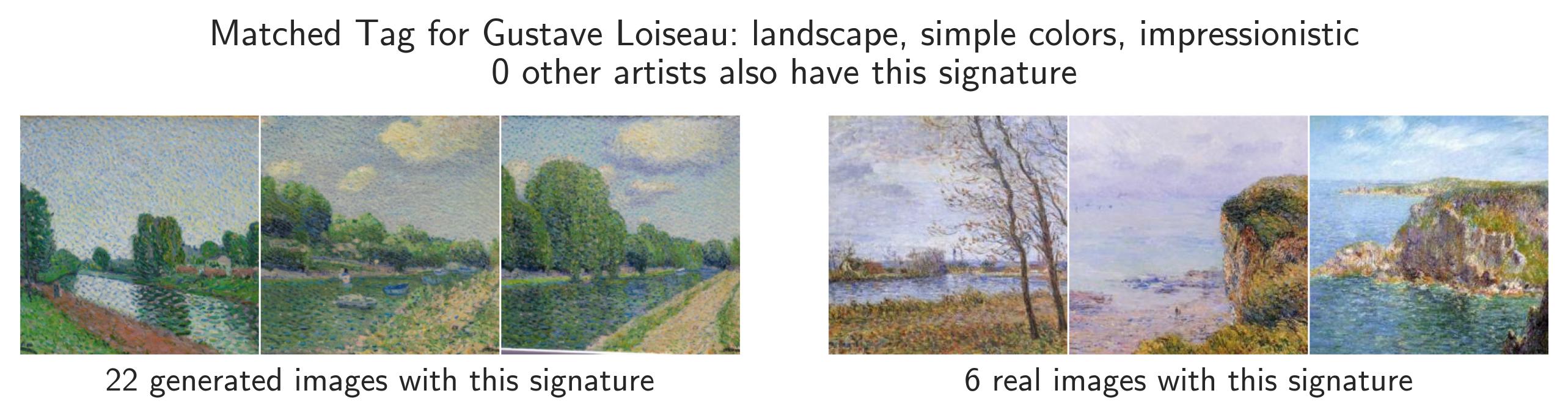}
    \includegraphics[width=\linewidth]{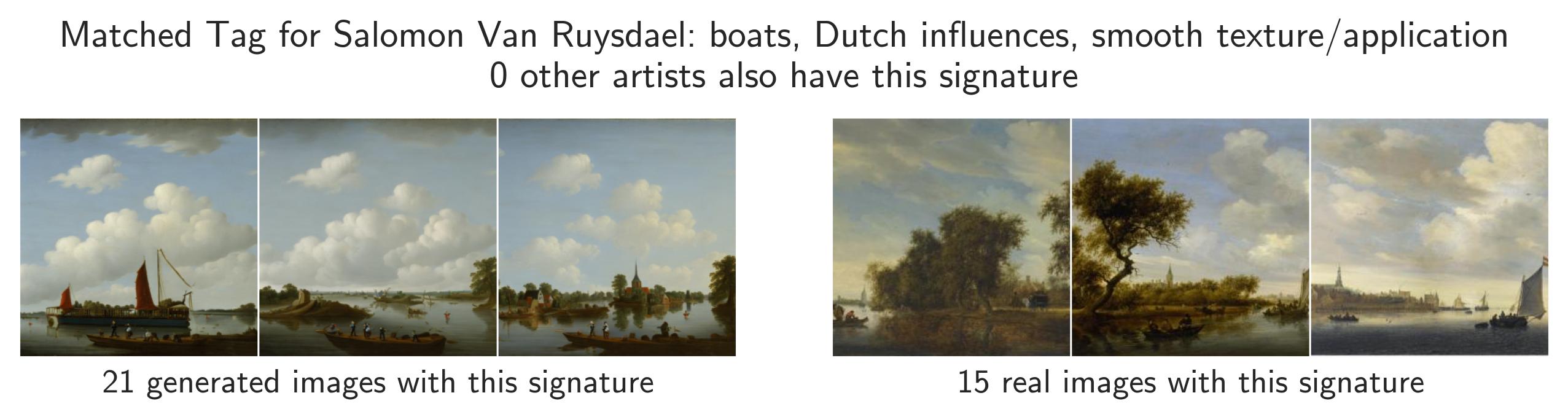}
    \includegraphics[width=\linewidth]{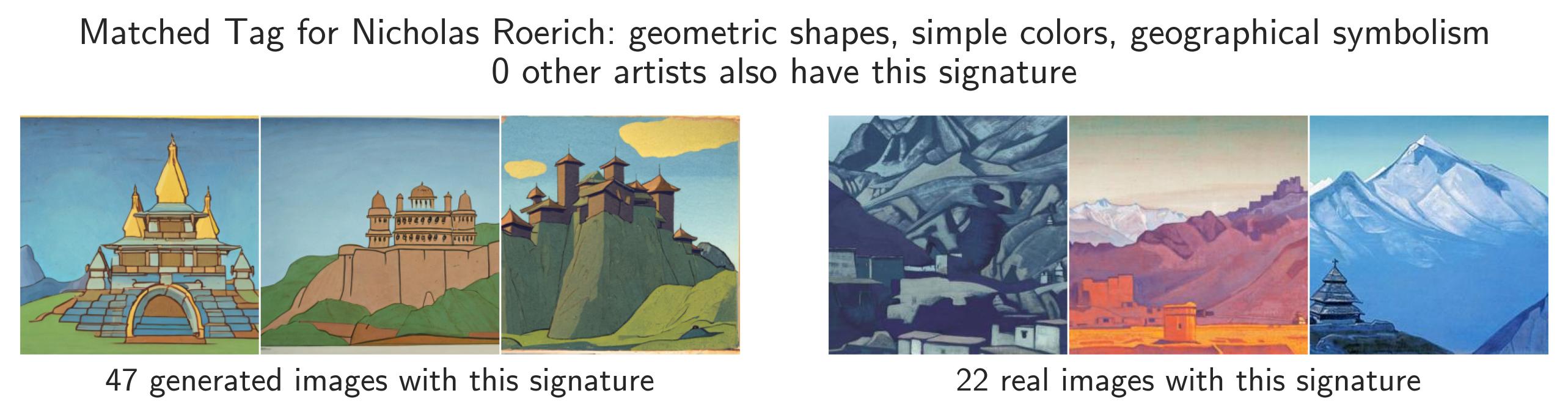}
    \caption{Examples of applying TagMatch to generated images. TagMatch is inherently interpretable with respect to tags, as each inference comes with the exact set of tags that are (i) shared between the sets of test art and art from the predicted artist (ii) used to predict the artist (i.e. the $10$ most unique shared tags; details in section \ref{sec: tagmatch}). Above, we visualize these matched tags for artists who's styles were recognized in their generated art.}
    \label{fig:tag_match}
    \vspace{-0.8cm}
\end{figure}

\looseness=-1
Figure \ref{fig:gen_imgs_artist_recog} shows match rate and match confidence across three generative models. We observe an average match rate of $20.2\%$, indicating that for the vast majority of artists in our study, \emph{generative models cannot reproduce their styles} in a way recognizable to DeepMatch, which has $89\%$ accuracy on real art. When we inspect match confidences (right subplot), we see that even for artists who's styles are recognized over a set of generations, the per-image accuracy is far lower. To be specific, averaged over models, about $50\%$ of artists see match confidences below $20\%$, indicating that less than one in five images generated in their style are predicted as their work. The mean confidence is $25.9\%$, and for all three models, more than half of the artists see match confidences below $20\%$. Indeed, $26\%$ of artists yield an average confidence below $5\%$. 
%Moreover, a surprising number of artists see match confidences of $0$, indicating that \emph{none} of the images generated in the styles of those artists match the intended artist's style, according to the the underlying classifier.  
On the other hand, we observe a handful of artists who's styles are matched with high confidence: $16$ artists see average match confidences of at least $75\%$. These include very famous artists like Van Gogh, Claude Monet, Renoir, which we'd expect generative models to do well in emulating. However, a few lesser known artists are also present, such as Cindy Sherman and Jacek Yerka, who are still alive, and thus could be negatively affected by generative models reproducing their styles. 
% \begin{figure}
%     \centering
%     \includegraphics[width=\linewidth]{figures/matched_tags_paper/AntoineBlanchard__SimpleColors_Streets_ContemporaryInfluences_SocialSymbolism.jpg}
%     \includegraphics[width=\linewidth]{figures/matched_tags_paper/FranzXaverWinterhalter__BroadBrushwork_FemaleFigures_HistoricalSymbolism.jpg}
%     \includegraphics[width=\linewidth]{figures/matched_tags_paper/GustaveLoiseau__Landscape_SimpleColors_Impressionistic.jpg}
%     \includegraphics[width=\linewidth]{figures/matched_tags_paper/SalomonVanRuysdael__Boats_DutchInfluences_SmoothTextureApplication.jpg}
%     \includegraphics[width=\linewidth]{figures/matched_tags_paper/NicholasRoerich__GeometricShapes_SimpleColors_GeographicalSymbolism.jpg}
%     % \includegraphics[width=\linewidth]{figures/matched_tags_paper/ArthurRackham__Illustration_Children_FantasticalSubjectMatter.jpg}
%     \caption{Examples of applying TagMatch to generated images. TagMatch is inherently interpretable with respect to tags, as each inference comes with the exact set of tags that are (i) shared between the sets of test art and art from the predicted artist (ii) used to predict the artist (i.e. the $10$ most unique shared tags; details in section \ref{sec:tagmatch}). Above, we visualize these matched tags for artists who's styles were recognized in their generated art.}
%     \label{fig:tag_match}
% \end{figure}

% \vspace{-0.2cm}
\subsection{Articulating Style Infringements with TagMatch}
We now utilize TagMatch because of its enhanced interpretability. Recall that in addition to predicting an artistic style, TagMatch also names the specific signature shared between the test set of images and the reference set of images for the predicted style. Thus, we can inspect the shared signature, as well as instances from both sets where the signature is present. Figure \ref{fig:tag_match} visualizes a number of examples. In comparing the subsets of generated and real images for top-1 matches, we observe that while pixel level differences are common across retrieved image subsets, stylistic elements are consistent in both sets with the labeled tags. Thus, TagMatch can serve as an effective tool for describing the way in which stylistic elements are copied via language, as well as providing direct evidence of the potential infringement. 

TagMatch yields match rates of $12.95\%$, $37.31\%$, and $52.3\%$ for top-1, top-5, and top-10 matches respectively. Low top-1 match rates using TagMatch alone cannot be used to argue that generative models do not reproduce artistic styles, but convincing arguments can be made combining the two methods. TagMatch also allows for understanding image distributions from the perspective of interpretable tags. We explore this direction in the appendix, finding differences in the uniqueness of the tags present in generated art vs real art. 

\section{Conclusion}
In our paper, we rethink the problem of copyright infringement in the context of artistic styles. We first argue that image-similarity approaches to copy detection may not fully capture the nuance of artistic style copying. After reformulating the task to a classification problem over image sets, we develop a novel tool -- \toolname, consisting of a dataset and two complementary methods that can effectively recognize artistic styles, via neural and tag based signatures. The success of our method offer strong evidence to the existence of unique artistic signatures, a necessary pre-requisite for styles to be protected. We highlight TagMatch, which scaffolds a black-box AI component with interpretable intermediate outputs and a transparent way in which intermediate outputs are combined to arrive at a final prediction, resulting in a white(r)-box AI \emph{system}. TagMatch, which can classify a set of images to an artist with reasonable accuracy as well as provide succinct text explanations and image attributions. Using these two detectors, we analyze generated images from different text-to-image generative models highlighting that amongst all the artists (including many famous ones), {\it only} 20$\%$ of the artists are recognized as having their style copied. 

% \clearpage\mbox{}Page \thepage\ of the manuscript.
% \clearpage\mbox{}Page \thepage\ of the manuscript.
% \clearpage\mbox{}Page \thepage\ of the manuscript.
% \clearpage\mbox{}Page \thepage\ of the manuscript.
% \clearpage\mbox{}Page \thepage\ of the manuscript. This is the last page.
% Now we have reached the maximum length of an ECCV \ECCVyear{} submission (excluding references).
% References should start immediately after the main text, but can continue past p.\ 14 if needed.
% \clearpage  % TODO REVIEW/FINAL: This \clearpage needs to be removed from both review and camera-ready versions.

\section{Acknowledgements}

This project was supported in part by a grant from an NSF CAREER AWARD 1942230, ONR YIP award N00014-22-1-2271, ARO’s Early Career Program Award 310902-00001, HR00112090132 (DARPA/ RED), HR001119S0026 (DARPA/ GARD), Army Grant No. W911NF2120076, the NSF award CCF2212458, NSF Award No. 2229885 (NSF Institute for Trustworthy AI in Law and Society, TRAILS), an Amazon Research Award and an award from Capital One.

\newpage
\bibliographystyle{splncs04}
\bibliography{main}

\begin{thebibliography}{10}
\providecommand{\url}[1]{\texttt{#1}}
\providecommand{\urlprefix}{URL }
\providecommand{\doi}[1]{https://doi.org/#1}

\bibitem{deepfloyd}
Deepfloyd (Apr 2023), \url{https://github.com/deep-floyd/IF}

\bibitem{CRS_Reports_2023}
Generative artificial intelligence and copyright law (Sep 2023), \url{https://crsreports.congress.gov/product/pdf/LSB/LSB10922}

\bibitem{basu2023localizing}
Basu, S., Zhao, N., Morariu, V., Feizi, S., Manjunatha, V.: Localizing and editing knowledge in text-to-image generative models (2023)

\bibitem{carlini2023extracting}
Carlini, N., Hayes, J., Nasr, M., Jagielski, M., Sehwag, V., Tramèr, F., Balle, B., Ippolito, D., Wallace, E.: Extracting training data from diffusion models (2023)

\bibitem{DBLP:journals/corr/abs-2104-14294}
Caron, M., Touvron, H., Misra, I., J{\'{e}}gou, H., Mairal, J., Bojanowski, P., Joulin, A.: Emerging properties in self-supervised vision transformers. CoRR  \textbf{abs/2104.14294} (2021), \url{https://arxiv.org/abs/2104.14294}

\bibitem{caron2021emerging}
Caron, M., Touvron, H., Misra, I., J{\'e}gou, H., Mairal, J., Bojanowski, P., Joulin, A.: Emerging properties in self-supervised vision transformers. In: Proceedings of the IEEE/CVF international conference on computer vision. pp. 9650--9660 (2021)

\bibitem{cui2024ftshield}
Cui, Y., Ren, J., Lin, Y., Xu, H., He, P., Xing, Y., Fan, W., Liu, H., Tang, J.: {FT}-{SHIELD}: A watermark against unauthorized fine-tuning in text-to-image diffusion models (2024), \url{https://openreview.net/forum?id=OQccFglTb5}

\bibitem{cui2023diffusionshield}
Cui, Y., Ren, J., Xu, H., He, P., Liu, H., Sun, L., Xing, Y., Tang, J.: Diffusionshield: A watermark for copyright protection against generative diffusion models (2023)

\bibitem{gandikota2023unified}
Gandikota, R., Orgad, H., Belinkov, Y., Materzyńska, J., Bau, D.: Unified concept editing in diffusion models (2023)

\bibitem{gatys2016image}
Gatys, L.A., Ecker, A.S., Bethge, M.: Image style transfer using convolutional neural networks. In: Proceedings of the IEEE conference on computer vision and pattern recognition. pp. 2414--2423 (2016)

\bibitem{goldsteincopyright}
Goldstein, P.: Goldstein on Copyright, 3rd edition. Wolters Kluwer Legal \& Regulatory U.S. (2014)

\bibitem{ram_plusplus}
Huang, X., Huang, Y.J., Zhang, Y., Tian, W., Feng, R., Zhang, Y., Xie, Y., Li, Y., Zhang, L.: Open-set image tagging with multi-grained text supervision. arXiv e-prints pp. arXiv--2310 (2023)

\bibitem{kumari2023ablating}
Kumari, N., Zhang, B., Wang, S.Y., Shechtman, E., Zhang, R., Zhu, J.Y.: Ablating concepts in text-to-image diffusion models (2023)

\bibitem{tufenkian}
Oakes, Calebrisi, Sotomayor: Tufenkian import export ventures inc v. einstein moomjy inc (2003), \url{https://caselaw.findlaw.com/court/us-2nd-circuit/1455682.html}

\bibitem{pizzi2022selfsupervised}
Pizzi, E., Roy, S.D., Ravindra, S.N., Goyal, P., Douze, M.: A self-supervised descriptor for image copy detection (2022)

\bibitem{podell2024sdxl}
Podell, D., English, Z., Lacey, K., Blattmann, A., Dockhorn, T., M{\"u}ller, J., Penna, J., Rombach, R.: {SDXL}: Improving latent diffusion models for high-resolution image synthesis. In: The Twelfth International Conference on Learning Representations (2024), \url{https://openreview.net/forum?id=di52zR8xgf}

\bibitem{radford2021learning}
Radford, A., Kim, J.W., Hallacy, C., Ramesh, A., Goh, G., Agarwal, S., Sastry, G., Askell, A., Mishkin, P., Clark, J., Krueger, G., Sutskever, I.: Learning transferable visual models from natural language supervision (2021)

\bibitem{ren2024copyright}
Ren, J., Xu, H., He, P., Cui, Y., Zeng, S., Zhang, J., Wen, H., Ding, J., Liu, H., Chang, Y., Tang, J.: Copyright protection in generative ai: A technical perspective (2024)

\bibitem{rezaei2023prime}
Rezaei, K., Saberi, M., Moayeri, M., Feizi, S.: Prime: Prioritizing interpretability in failure mode extraction (2023)

\bibitem{rombach2021highresolution}
Rombach, R., Blattmann, A., Lorenz, D., Esser, P., Ommer, B.: High-resolution image synthesis with latent diffusion models (2021)

\bibitem{saharia2022photorealistic}
Saharia, C., Chan, W., Saxena, S., Li, L., Whang, J., Denton, E., Ghasemipour, S.K.S., Ayan, B.K., Mahdavi, S.S., Lopes, R.G., Salimans, T., Ho, J., Fleet, D.J., Norouzi, M.: Photorealistic text-to-image diffusion models with deep language understanding (2022)

\bibitem{schuhmann2022laion5b}
Schuhmann, C., Beaumont, R., Vencu, R., Gordon, C., Wightman, R., Cherti, M., Coombes, T., Katta, A., Mullis, C., Wortsman, M., Schramowski, P., Kundurthy, S., Crowson, K., Schmidt, L., Kaczmarczyk, R., Jitsev, J.: Laion-5b: An open large-scale dataset for training next generation image-text models (2022)

\bibitem{schuhmann2022laionb}
Schuhmann, C., Beaumont, R., Vencu, R., Gordon, C.W., Wightman, R., Cherti, M., Coombes, T., Katta, A., Mullis, C., Wortsman, M., Schramowski, P., Kundurthy, S.R., Crowson, K., Schmidt, L., Kaczmarczyk, R., Jitsev, J.: {LAION}-5b: An open large-scale dataset for training next generation image-text models. In: Thirty-sixth Conference on Neural Information Processing Systems Datasets and Benchmarks Track (2022), \url{https://openreview.net/forum?id=M3Y74vmsMcY}

\bibitem{shan2023glaze}
Shan, S., Cryan, J., Wenger, E., Zheng, H., Hanocka, R., Zhao, B.Y.: Glaze: Protecting artists from style mimicry by text-to-image models (2023)

\bibitem{somepalli2022diffusion}
Somepalli, G., Singla, V., Goldblum, M., Geiping, J., Goldstein, T.: Diffusion art or digital forgery? investigating data replication in diffusion models (2022)

\bibitem{NEURIPS2023_9521b6e7}
Somepalli, G., Singla, V., Goldblum, M., Geiping, J., Goldstein, T.: Understanding and mitigating copying in diffusion models. In: Oh, A., Neumann, T., Globerson, A., Saenko, K., Hardt, M., Levine, S. (eds.) Advances in Neural Information Processing Systems. vol.~36, pp. 47783--47803. Curran Associates, Inc. (2023), \url{https://proceedings.neurips.cc/paper_files/paper/2023/file/9521b6e7f33e039e7d92e23f5e37bbf4-Paper-Conference.pdf}

\bibitem{DBLP:journals/corr/TanCAT17}
Tan, W.R., Chan, C.S., Aguirre, H.E., Tanaka, K.: Artgan: Artwork synthesis with conditional categorial gans. CoRR  \textbf{abs/1702.03410} (2017), \url{http://arxiv.org/abs/1702.03410}

\bibitem{wang2024diagnosis}
Wang, Z., Chen, C., Lyu, L., Metaxas, D.N., Ma, S.: {DIAGNOSIS}: Detecting unauthorized data usages in text-to-image diffusion models. In: The Twelfth International Conference on Learning Representations (2024), \url{https://openreview.net/forum?id=f8S3aLm0Vp}

\bibitem{xue2024effective}
Xue, H., Liang, C., Wu, X., Chen, Y.: Toward effective protection against diffusion based mimicry through score distillation (2024)

\bibitem{zhao2023protective}
Zhao, Z., Duan, J., Xu, K., Wang, C., Guo, R.Z.Z.D.Q., Hu, X.: Can protective perturbation safeguard personal data from being exploited by stable diffusion? (2023)

\bibitem{zheng2023judging}
Zheng, L., Chiang, W.L., Sheng, Y., Zhuang, S., Wu, Z., Zhuang, Y., Lin, Z., Li, Z., Li, D., Xing, E., Zhang, H., Gonzalez, J.E., Stoica, I.: Judging {LLM}-as-a-judge with {MT}-bench and chatbot arena. In: Thirty-seventh Conference on Neural Information Processing Systems Datasets and Benchmarks Track (2023), \url{https://openreview.net/forum?id=uccHPGDlao}

\end{thebibliography}
\newpage

\appendix

\section{\toolname\; Demonstration:\\A Practical Tool to Protect Artists}
\begin{figure}[t]
    \centering
    \includegraphics[width=\linewidth]{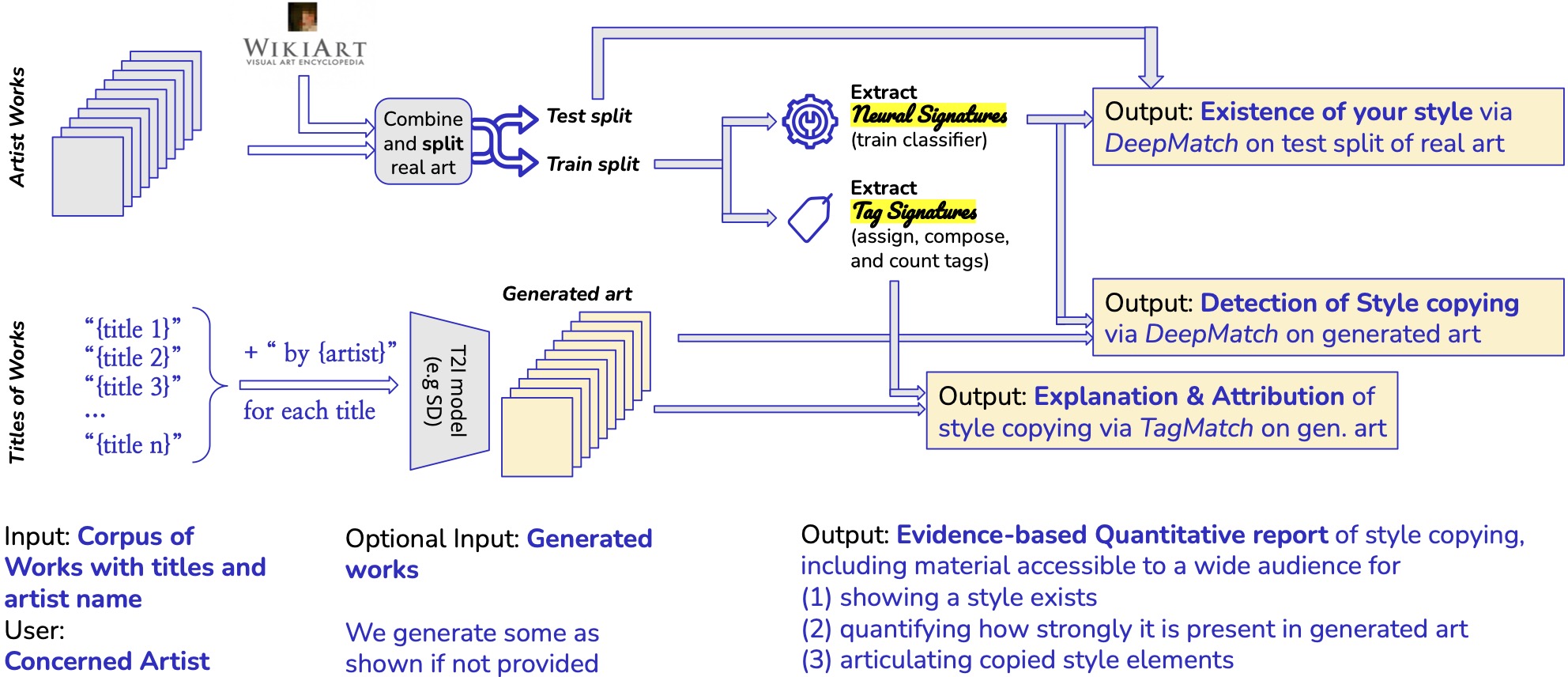}
    \caption{General flow demonstrating usage of \toolname. We design our tool with a concerned artist in mind, who wishes to quickly investigate the degree to which they may be at risk of style copying by generative AI models.}
    \label{fig:flow}
\end{figure}

The founding principle that \toolname\; was \emph{practical utility}. In other words, we strove to build something people can actually use. Specifically, the motivating use case revolved around a hypothetical artist who is concerned with generative models potentially copying their styles. In figure \ref{fig:flow}, we outline the general flow of how our tool can be used. The concerned artist would first present a corpus of their works, along with their own name and the titles of each work. Then, \toolname\; would create an easy-to-understand report characterizing the degree to which generative models copy the styles of the artist. The artist can present a set of generated images, or we can generate them by prompting text-to-image models with captions of the form ``\{title of work\} by \{name of artist\}'' for each work provided by the artist. 

As explained in the diagram, we first combine the provided works with our existing art repository, performing a train/test split as well. Using the train split, we extract neural and tag signatures. In other words, we train a classifier over the $372+1$ artists, and we also tag all images, compose tags within artists, and store extracted tags per artist. Then, using the extracted neural and tag signatures, we can apply DeepMatch and TagMatch respectively. Applying DeepMatch to the held-out art provides a measure of recognizability. That is, it establishes that the test artist has an identifiable style to begin with; otherwise, there can be no style copying. Then, running DeepMatch on generated images provides a quantitative manner to understand if the artist's style appears consistently in generated works (i.e. is there a match), and to what frequency (i.e. what is the match confidence).  Finally, running TagMatch on the generated images helps articulate the particular style signatures that are copied, enabling an analytic way to argue infringement. Moreover, TagMatch surfaces stylistically similar examples in a way that is faithful to how TagMatch infers styles.

\begin{figure}[t]
    \centering
    \includegraphics[width=\linewidth]{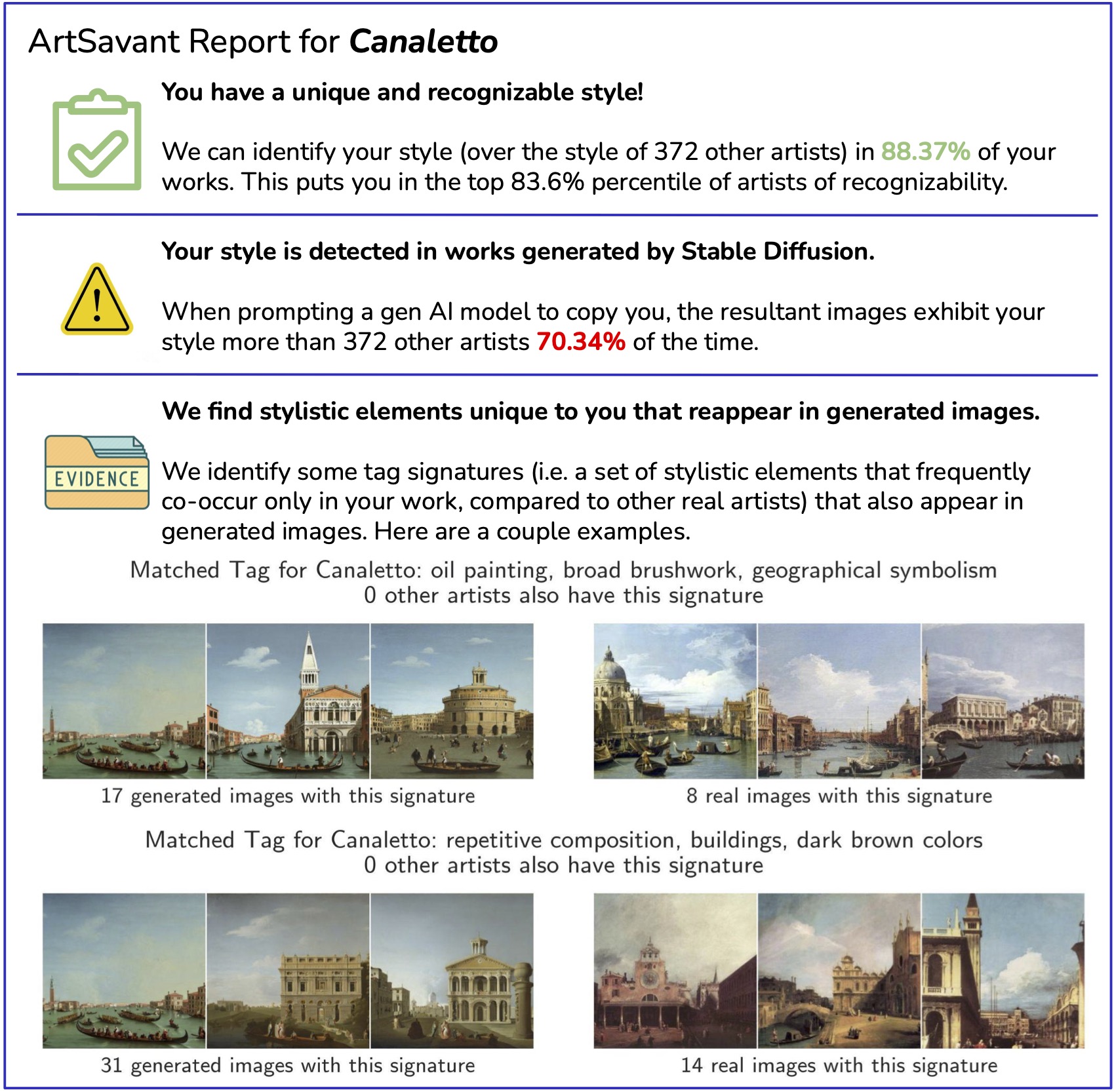}
    \caption{An example output report produced by ArtSavant. Here, when inspecting art from Canaletto, ArtSavant identifies a unique style and observes via both DeepMatch and TagMatch that a generative model replicates Canaletto's style too.}
    \label{fig:eg_report}
\end{figure}

Figure \ref{fig:eg_report} shows an example report outputted by \toolname when presented with art from an artist named Canaletto, who we observed was at risk of style infringement. We design the report to be easy to read and understand, as well as being based on evidence. Moreover, the report can be generated very quickly. Because all steps operate on embeddings from a frozen CLIP encoder, the process takes about $1-2$ minutes, as we can simply compute embeddings once (and offline for the WikiArt corpus). Note that our empirical analysis suggests that most artists are not at risk of style copying. While we do not provide an example output report, it would primarily consist of the first two components, and hopefully relieve concerned artists, by showing that the generated works fail to recognizably capture their unique styles. 

\section{Limitations}

Our work tackles a novel problem of artistic \emph{style} infringements. Style, however, is qualitative. We merely put forward one definition for artistic style, along with two implementations for demonstrating the existence of a style given example works from an artist and recognizing the identified style in other works. 

Importantly, we argue that an artist's style is unique if we can consistently distinguish their work from that of other artists. However, we can only proxy the entire space of artists. We construct a dataset consisting of works from $372$ artists spanning diverse schools of art and time periods in attempt to represent the space of existing artists, though of course we will always fall short in capturing all kinds of art. We provide tools to allow for this dataset to grow with time, and we caution that if only one artist for some broader artistic style is not present in our reference set, the uniqueness of that artist's style may be overestimated, and as such, generated images may be matched to this artist with an overestimated confidence. However, if only one out of $372$ artists exhibits some style, than one could argue that that alone reflects a notable uniqueness of that artist. To employ a stricter criterion for alleging style copying,  we'd recommend augmenting the reference set to include more artists with very similar styles to the artist in question. Nonetheless, we believe our reference dataset does well in representing all art, to where analysis based on this reference set is still informative.

We also note that our atomic tagging leverages an existing foundation model (CLIP) with no additional training. While we verify the precision of our tags, CLIP is known to have issues with complex concepts. Further, we do not claim our tags achieve perfect recall (most image taggers do not). We advise users to interpret the assignment of a tag to indicate a strong presence of that concept, relative to similar concepts (i.e. from the same aspect of artistic style). While our tagger is not perfect, it is objective and automatic, enabling interpretable style articulation and detection. Also, we note that the field of image tagging in general has seen rapid improvement in the past year \cite{ram_plusplus}, and an improved tagger could easily be swapped into our pipeline.
% TODO: discuss hyperparams? Mainly the absolute threshold for tag count thresholds, may lead to 

Lastly, we only analyze generated images using off-the-shelf text-to-image models. It is possible that particularly determined and AI-adept style thiefs fine-tune a model to more closely replicate specific artistic styles. This is a much more threatening scenario, though requires greater effort and ability by the style thief. We elect to demonstrate the feasability of our approach in the more broadly accessible setting of using models off-the-shelf, and note that our method can flexibly accept generated images produced in a different way (or perhaps discovered on the internet); notice generated images are an optional input in figure \ref{fig:flow}. We look forward to explorations of more threatening scenarios in future work, and hope both our formulation and methods for measuring style copying prove to be of use.

\section{A nuance in artistic style infringements:\\Existing Artists can have very similar styles}

A crucial step in arguing that an artist's style has been infringed is to first demonstrate the existence of the given artist's \emph{unique} style. We note that doing so objectively is non-trivial, as a style may not have a clear definition, and thus, it can be challenging to systematically compare to all other artistic styles, so to show uniqueness. In our work, we utilized classification, claiming that if an artist's works can consistently be mapped (i.e. at least half the time) to that artist (over a large set of other artists), than that artist must have some underlying unique style (parameterized by a neural signature). 

In doing so, we found that $89.3\%$ of artists could be recognized based of a set of (at least $20$ of) their works (held-out in training the classifier). What about the remaining $10.7\%$ of artists? We now take a closer look at these artists, and also introduce a second, stricter style copying criterion. Namely, we consider the notion that it may be unfair to claim a generative model is copying the style of an artist, if another existing artist seems to also be copying that artist. That is, we propose a way to verify that the generative model not only shows a substantial similarity to the copied artist, but also an \emph{unprecedented} similarity.

\begin{figure}[t]
    \centering
    \includegraphics[width=0.325\linewidth]{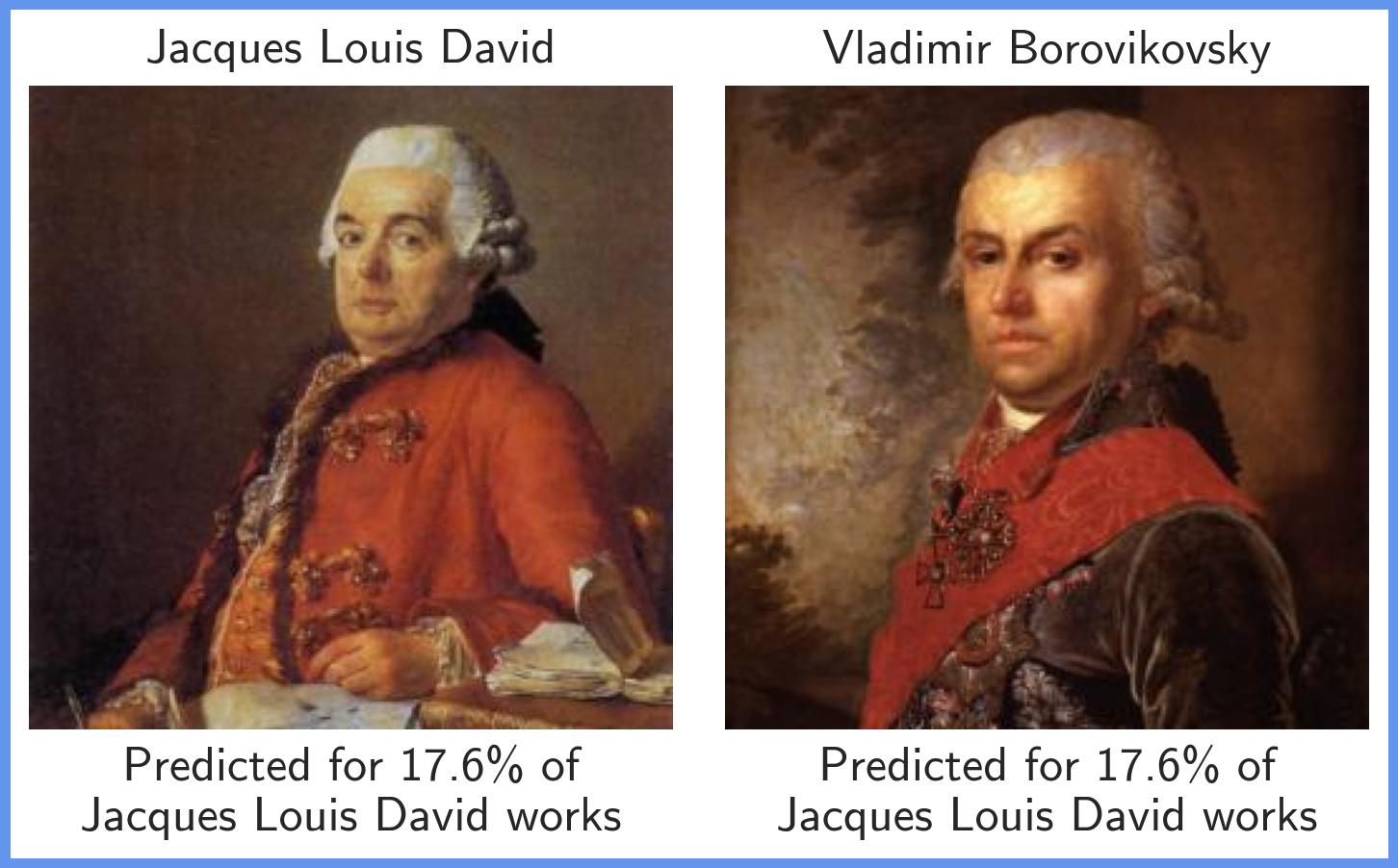}
    \includegraphics[width=0.325\linewidth]{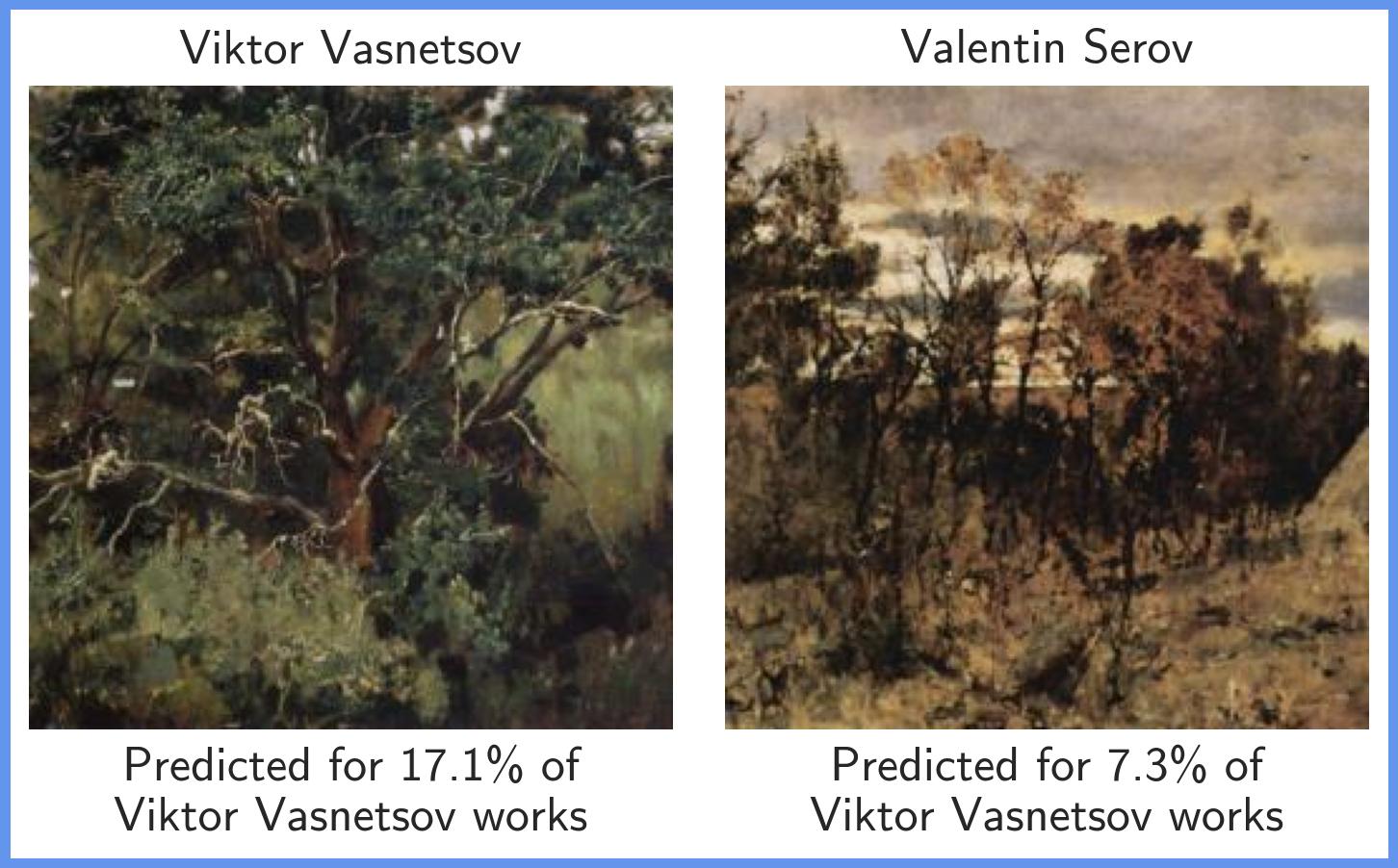}
    \includegraphics[width=0.325\linewidth]{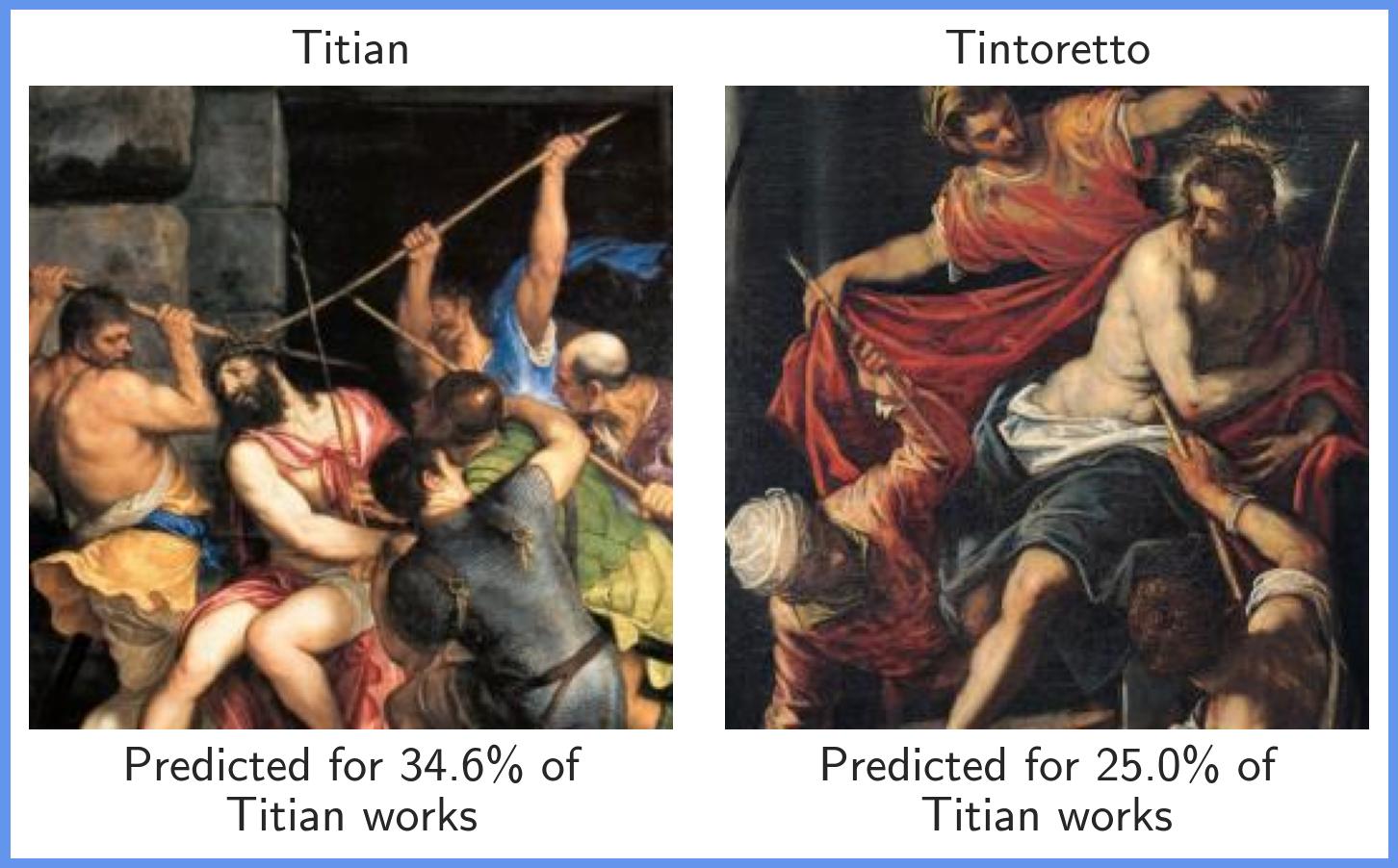}
    \includegraphics[width=0.325\linewidth]{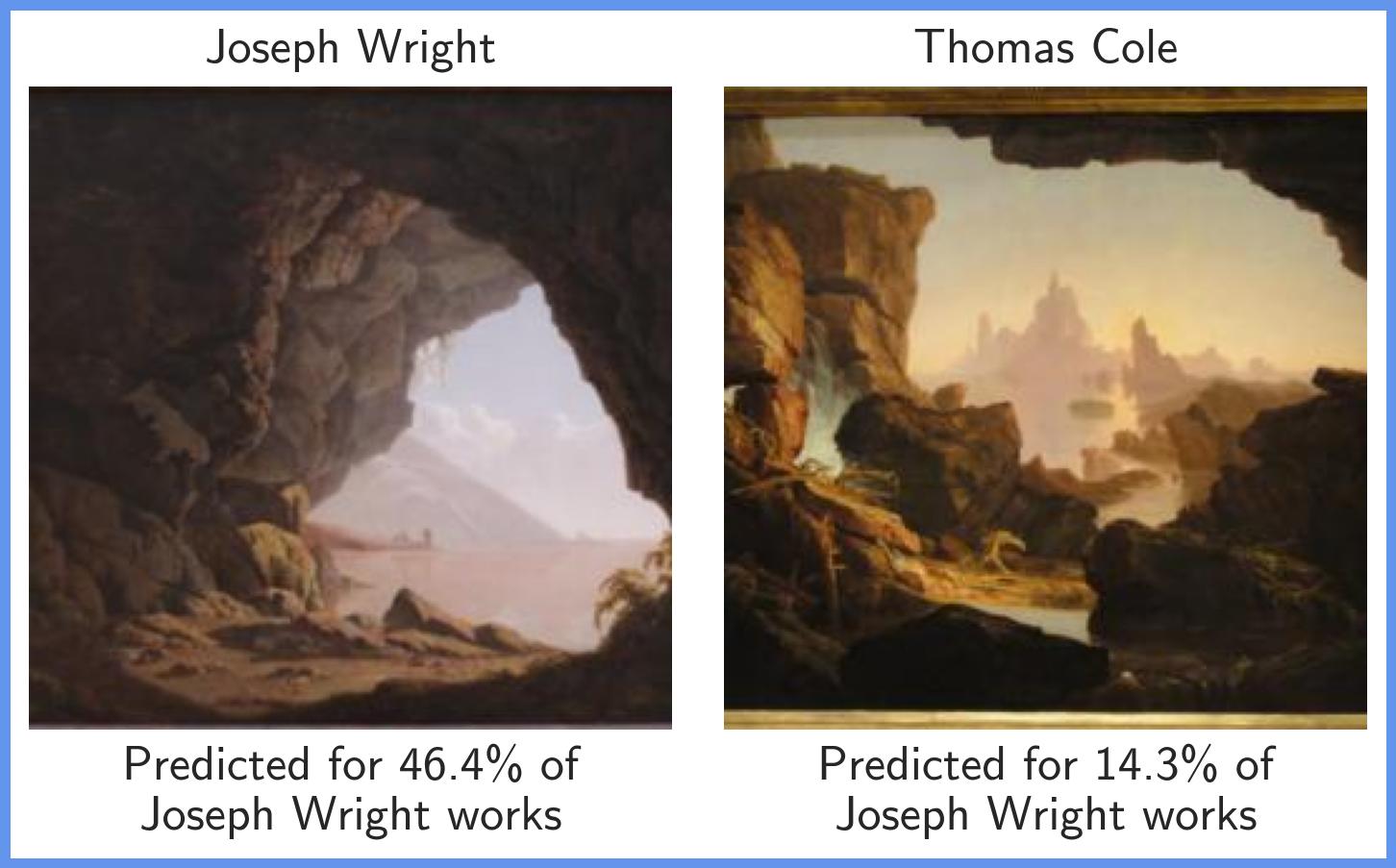}
    \includegraphics[width=0.325\linewidth]{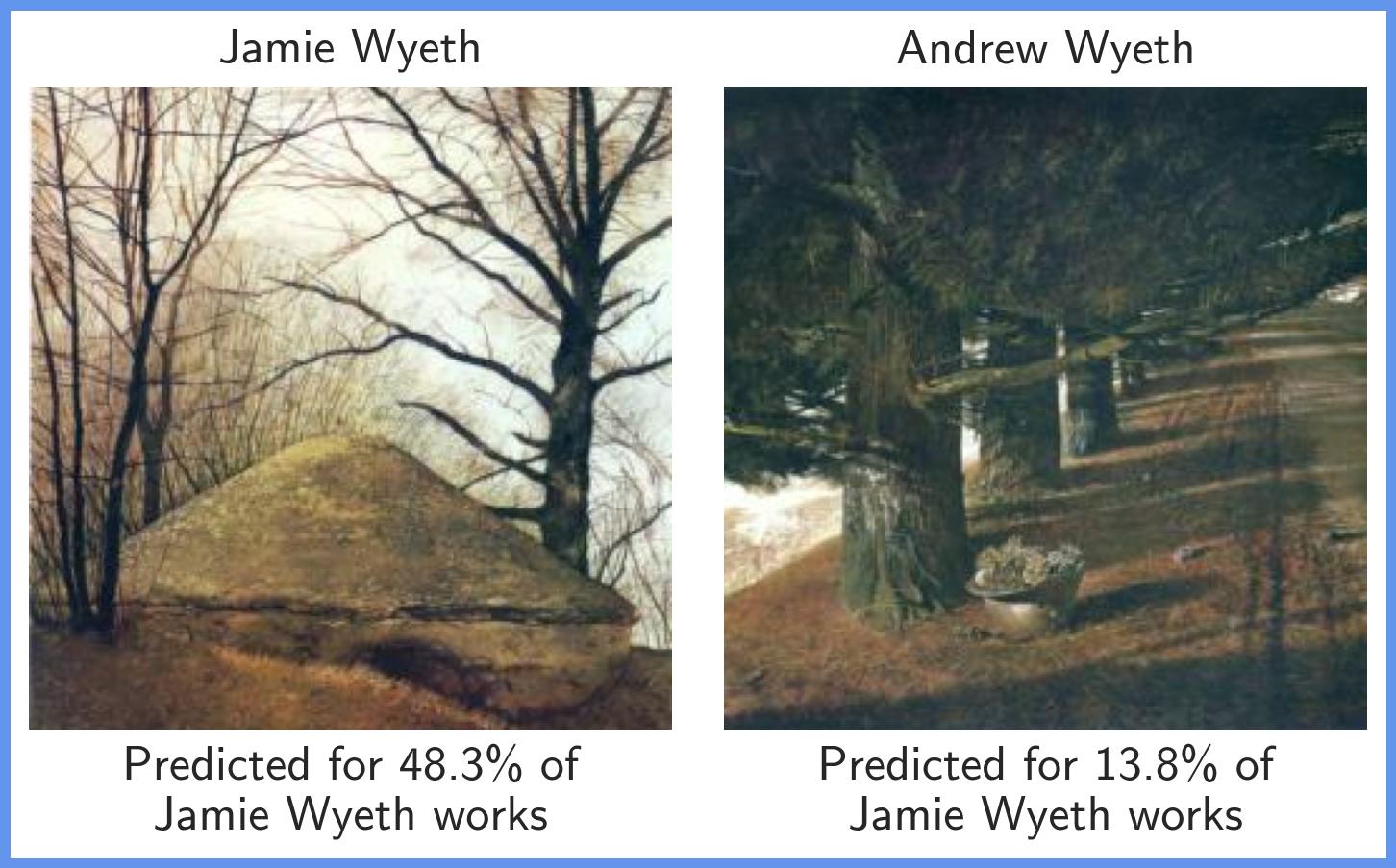}
    \includegraphics[width=0.325\linewidth]{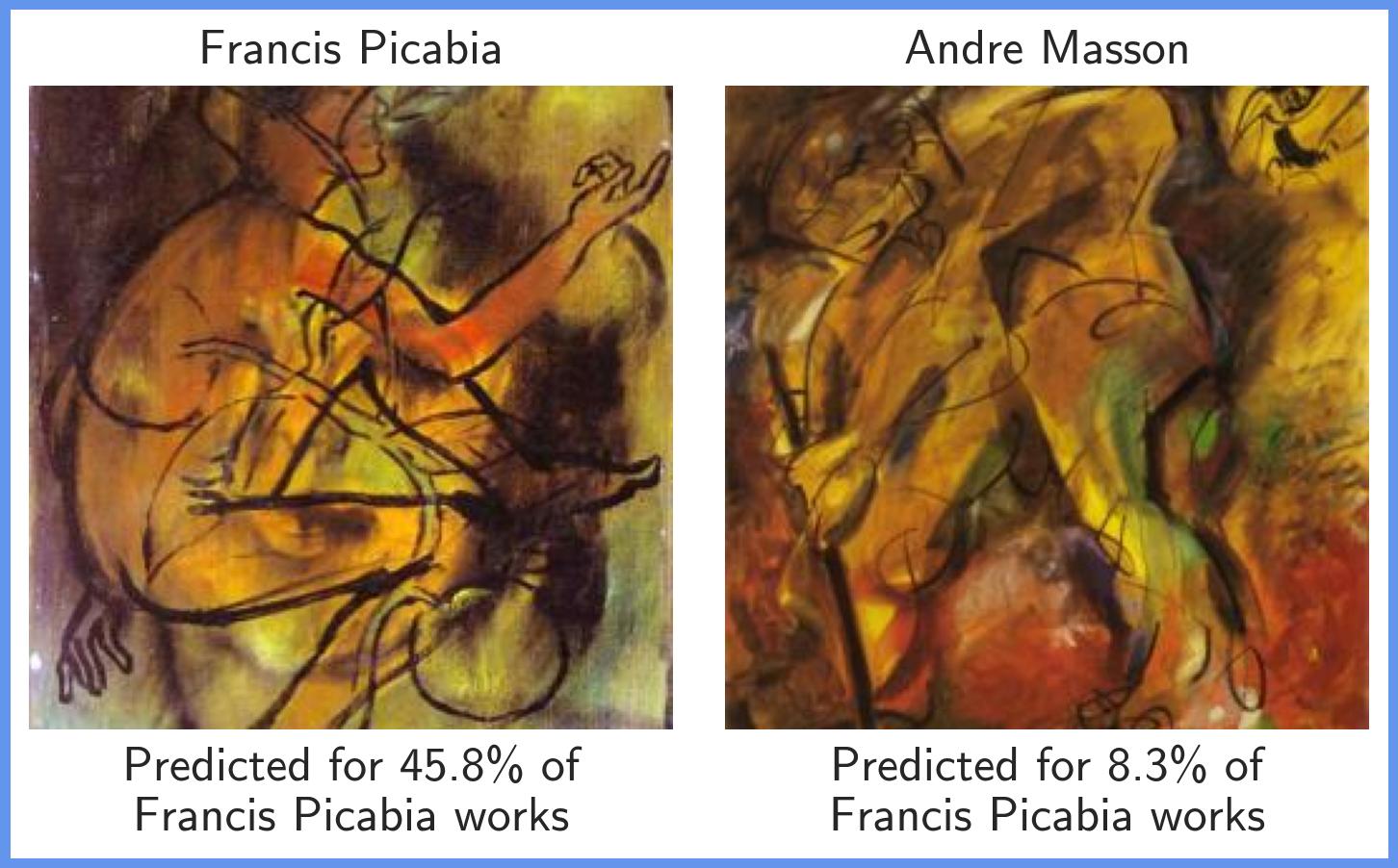}
    \includegraphics[width=0.325\linewidth]{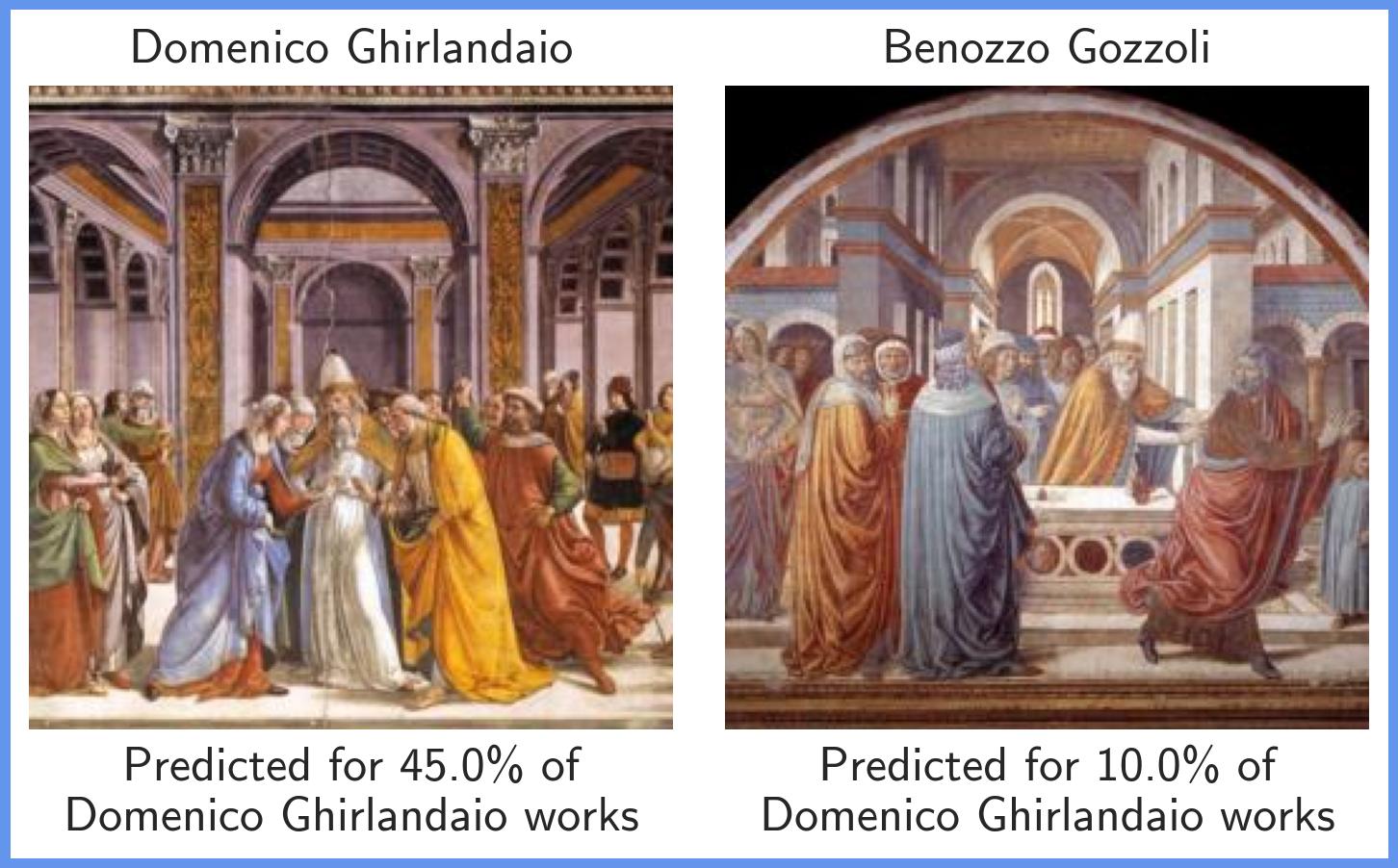}
    \includegraphics[width=0.325\linewidth]{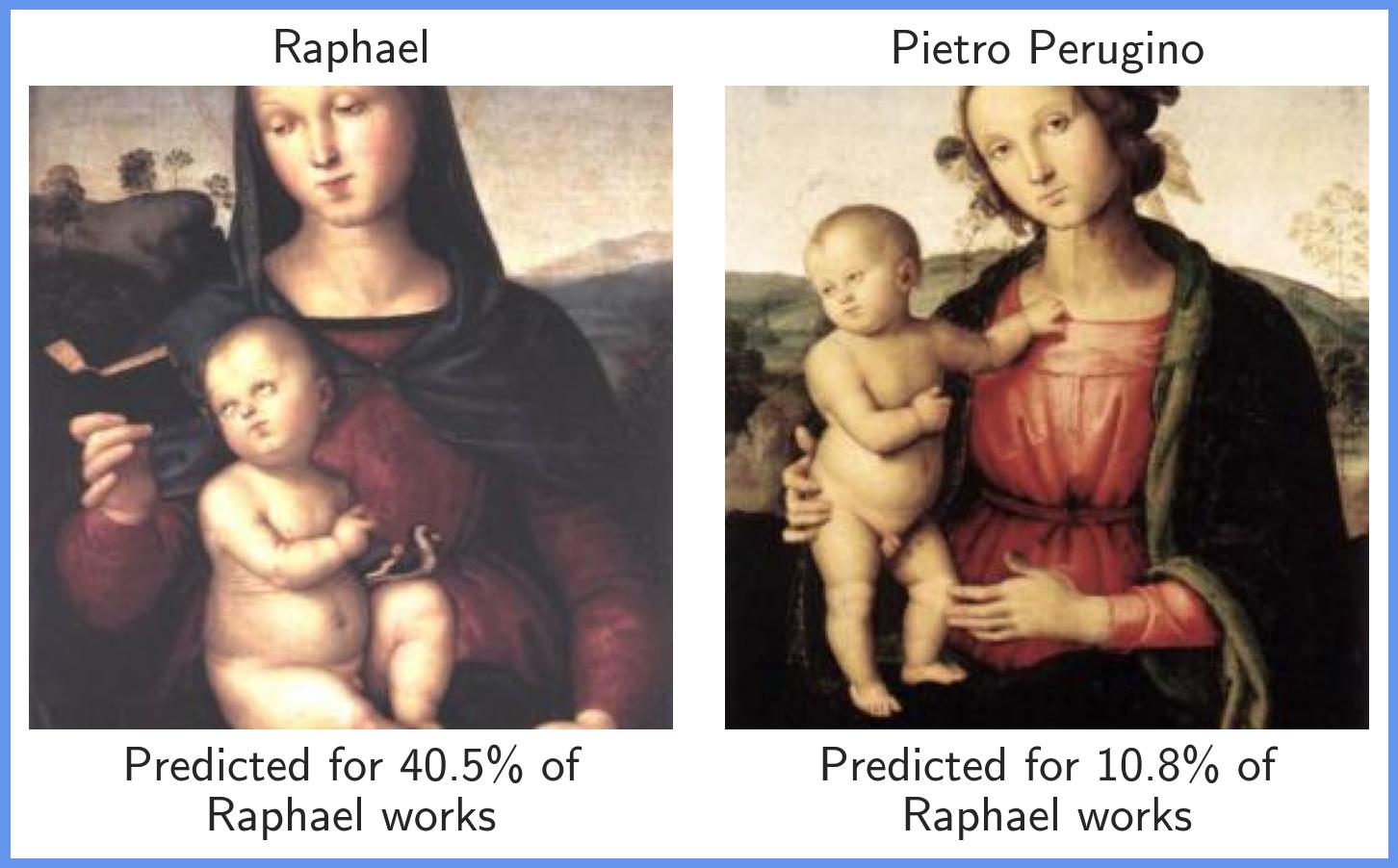}
    \includegraphics[width=0.325\linewidth]{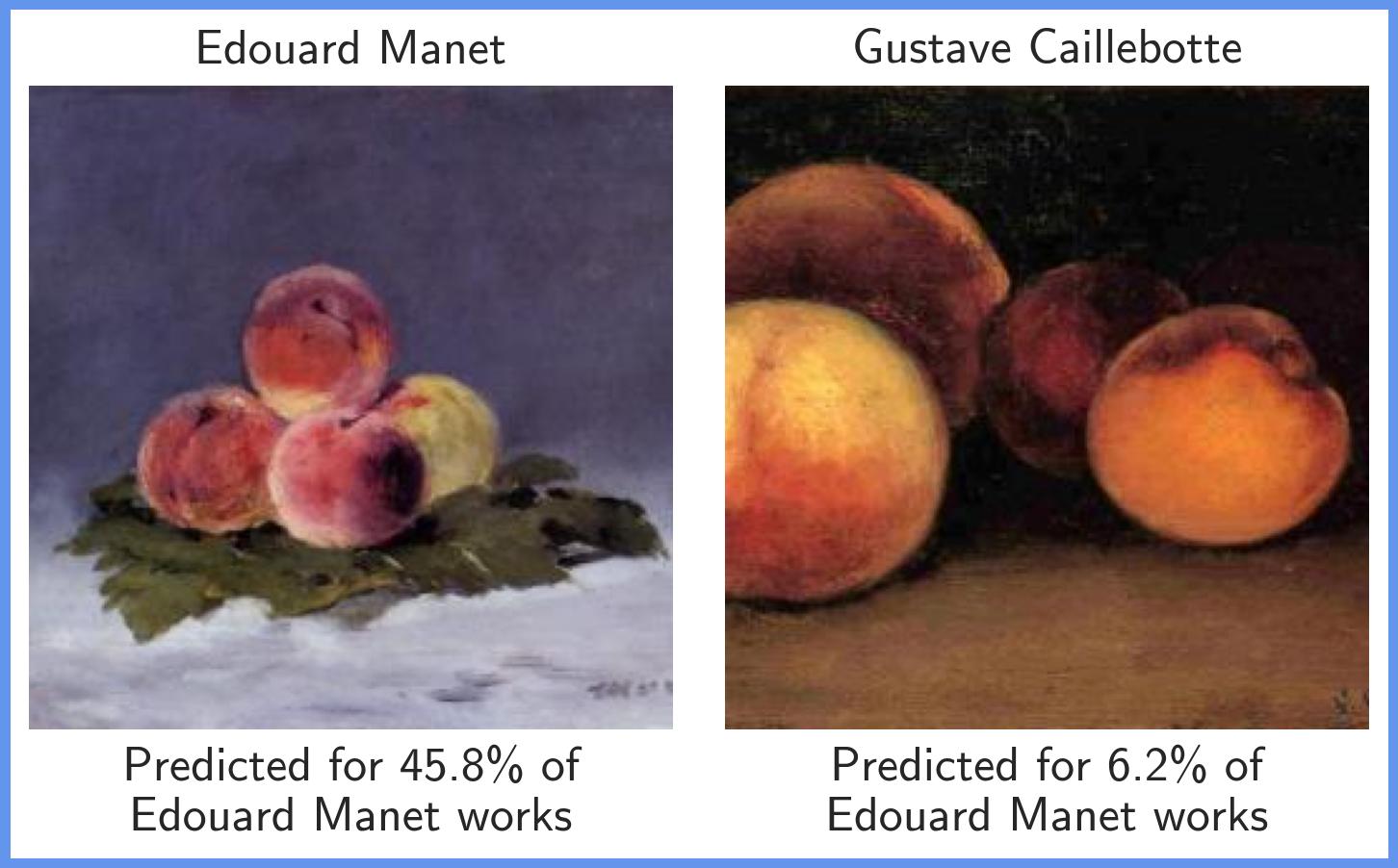}
    \caption{Examples of artists who's styles were not recognized by DeepMatch (i.e. less than half of their held-out works were predicted to the artist). Each panel shows an example work from (left) the unrecognized artist and (right) the artist that is incorrectly predicted most frequently over works from the unrecognized artist. We see that artists can use very similar, at times arguably indistinguishable, styles.}
    \label{app-fig:unrecog_artists}
\end{figure}

\subsection{Artists who's styles were not recognized}

First, we inspect more examples from artists who were not recognized using our majority voting threshold in DeepMatch. That is, less than half of their held-out works were predicted to them. Figure \ref{app-fig:unrecog_artists} shows a number of examples, from which we can make some qualitative observations. First, the styles of artists who operate in the same broader genre (e.g. portraiture, landscapes, narrative scenes in renaissance styles, etc) can be extremely similar. We even see an instance where an artist's son's style is indistinguishable from his father's (Jamie and Andrew Wyeth). Lastly, we note that in most cases, the artists only marginally fall short of our recognition threshold (i.e. accuracy for their held-out works is only a bit below $50\%$). We utilize majority voting because (i) it is intuitive, (ii) it requires \emph{consistent} appearance of the neural signature across works, and (iii) it allows for abstention when no particular style is strongly present. However, the exact threshold of $50\%$ can be altered as desired. In summary, as in Figure \ref{fig:unrecog_artists}, we see artistic styles can be very similar, making the existence of unique artistic styles for the vast majority of artists a non-trivial observation. 

If an artist's style cannot be recognized over their own held-out works, arguing that a generative model copies that style is strenuous, as the style itself is ill-defined. Notably, in these cases, the classifier had an option to predict the correct artist. However, in applying DeepMatch to generated images, there is no direct option for the classifier to abstain from predicting anyone, under that generated art comes from a ``new artist'', which takes inspiration from existing artists. Note that abstention is still possible (due to the majority voting in DeepMatch), and occurs when a match confidence falls below $50\%$. To make comparisons fairer to generative models, we now discuss a stricter criterion of \emph{unprecedented similarity}.  

\subsection{\emph{Unprecedented Similarity}: Do generative models copy styles more than existing artists already do?}

A nuance that requires consideration when studying artistic style copying is that it is possible for two artists to have very similar styles. Thus, it may be unfair to allege that a generative model is copying an artist $a$ if there exists another artist $b$ who's style is just as or in fact even more similar to artist $a$. Towards this end, we introduce \emph{unprecedented similarity}, which requires that the similarity between works of a generative model $A'$ and works of the artist inteded to be copied $A$ is higher than the similarity of any existing artist with $A$. That is, $sim(A, A') \geq sim(A, B)$ for works $B$ from all other existing artists $b$. 

\begin{figure}[t]
    \centering
    \includegraphics[width=\linewidth]{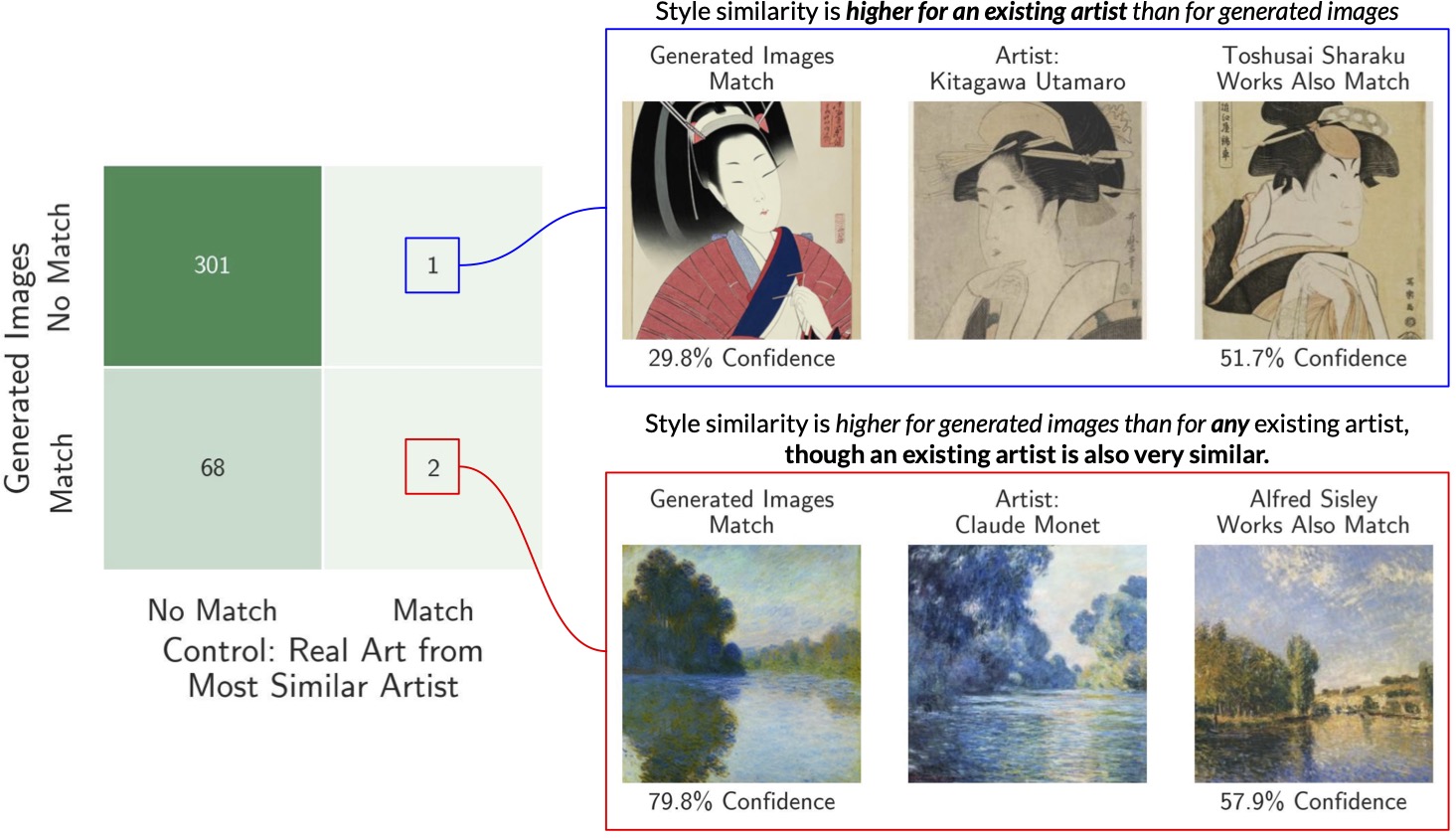}
    \caption{We verify the stricter criterion of \emph{unprecedented similarity} by holding out the real artist with highest similarity to a given artist, and checking if the held-out real artist's works are flagged as potential style copying by DeepMatch. ({\bf left}) We observe only three artists where the most similar held-out artist has their work flagged as a style match, and in all cases, when generated images are flagged, the match confidence of the generated images exceeds that of the held-out real artist's works (i.e., {\bf the generated images flagged by our method reflect \emph{unprecedented similarity} to the given artist's style}). ({\bf right}) Inspecting the flagged held-out artists further show that style copying is very nuanced, as artists take inspiration from one another, and as such, they may already have very similar styles. While we always observe unprecedented similarity, a potential solution to style copying may be for generative models to ensure that they do not copy any more than what already exists; that is, they may exhibit some copying, but no more than for which precedent already exists.}
    \label{fig:unprecedented_sim}
\end{figure}

Note that this is a stricter criterion than our previous threshold. In DeepMatch, we required that at least half of the works in a given set of test images were predicted to a single artist in order for us to flag the test images as a potential style infringmenet. In other words, that threshold required that $sim(A, A') \geq 0.5$, which in turn implies that $sim(A, A') \geq sim(A', B)$ for all $B$ (with room to spare; here we use match confidence to denote similarity). 

Now, however, instead of just comparing $A'$ to all $B$, we must also compare all $B$ to $A$. Instead of comparing all other artists, we inspect the most similar artist $b^*$ to $a$, identified by taking the artist $b$ with the highest rate of false positive predictions to artist $a$. Then, we hold out $b$, and train a new classifier on the remaining $371$ artists. Finally, we check for style matches of for the set of  generated images $A'$ and the works $B^*$ from the most similar artist $b^*$. 

Figure \ref{fig:unprecedented_sim} summarizes our result for OpenJourney (all three models studied show consistent results). We find that only in three cases do we see a held-out artist's work flagged as potential style copying. Notably, in all instances where generated work is flagged as potential style copying, the corresponding held-out artist's work is either not flagged or is flagged with lower confidence, indicating that the instances of style copying of generative models that we observe always also satisfy the criterion of unprecedented similarity. 

Taking a closer look at instances where held-out art is flagged for style copying (or perhaps style emulation?), we again see just how similar the works of different artists can be. Namely, we see that some artists works seem to fall into a broader genre of art that many artists utilize (e.g. ukiyo-e or impressionism). In summary, while generative models can very closely resemble the style of a given artist, contextualizing copying by generative models with respect to copying (or perhaps, `style emulation') already done by existing artists is crucial in order to afford the same artistic liberties to generative models as have been provided to other artists in the past.

\section{Details on TagMatch}

We now provide greater details regarding the implementation of TagMatch, a central technical contribution of our work. TagMatch is a method to classify a set of images to a class; specifically, we map a set of artworks to one artist, selected over $372$ choices. TagMatch is not as accurate as DeepMatch, as it maps held-out works of each artist in our WikiArt dataset to the correct artist about $61\%$ of the time (compared to $89\%$ top-1 accuracy for DeepMatch). However, top-5 accuracy is more reasonabe, achieving above $80\%$. Most notably, {\bf TagMatch is inherently interpretable and attributable}. It consists of three steps: (i) assigning atomic tags to images, (ii) efficiently composing tags to obtain more unique \emph{tag signatures}, and (iii) matching a test set of images to a reference artist based on the uniqueness of the tags shared between the test set and works from the predicted reference artist.

Our method is fast and flexible: after caching image embeddings, the whole thing only takes minutes, and it is easy to modify the concept vocabulary as desired, as the tagging is done in a zero-shot manner. Through MTurk studies, we verify that the atomic tags we assign our mostly precise, though we recognize that these descriptors can be subjective. Thus, while we do not claim perfect tagging, we stress that our method is easy to understand, and crucially, is deterministic per image. Therefore, ideally our tagging may be more reliable biased than human judgements, particularly when the humans involved may be biased (e.g. an artist alleging copying and a lawyer defending a generative model would have strong and opposing stakes).  

Below, we provide details for image tagging (\S \ref{app-sec:atomic_tags}), artist tagging (\S \ref{app-sec:tag_comp}), artistic style inference via tag matching (\S \ref{app-sec:tag_match_inference}), effect of hyperparameters (\S \ref{app-sec:hyperparams}), details on efficiency (\S \ref{app-sec:time_for_tagmatch}), and a review of validation (\S \ref{app-sec:tagmatch_val}).

\subsection{Image Tagging}
\label{app-sec:atomic_tags}

As explained in \S \ref{subsec:tagging}, we utilize CLIP to attain a diverse set of atomic tags per image in a zero-shot manner. Specifically, we first define a vocabulary of descriptors along various aspects of artistic style. Then, given an image, we do selective multi-label zero-shot classification \emph{for each aspect}. Performing zero-shot classification per aspect proves to be critical in order to achieve a diversity of tags and a similar number of tags per image. We find that some descriptors always lead to higher CLIP similarities than others. Specifically, descriptors for simple aspects, like colors and shapes, yield higher similarities than more complex aspects like brushwork and style. Thus, using a global threshold across descriptors would lead to a less diverse descriptor set. Moreover, we observe some images have higher similarities across the board than others, which again would lead global thresholding to result in a disparate number of tags per image. Our per-aspect scheme requires that the descriptors within each aspect are mostly mutually exclusive; we prioritize this in the construction of the concept vocabulary, via the prompt we present the LLM assistants and our manual verification. 

Namely, we prompt both Vicuna-33b and ChatGPT with ``{\it I want to build a vocabulary of tags to be able to describe art. First, consider different aspects of art, and then for each aspect, list about 20 distinct descriptors that could describe that aspect of art. Please return your answer in the form of a python dictionary.} ''. We then perform a filtering step with a human in the loop, where we manually remove tags that are difficult to recognize or redundant. After this filtering step, we add in a few new aspects. First, we incorporate the $20$ {\it styles} (e.g., ``impressionism'') and {\it genres} (e.g., ``portrait'') that are most common amongst works in our WikiArt dataset; note that all WikiArt images also contain metadata for these categories. Finally, we add some easy to understand tags such as {\it color} and {\it shape} which can be important characteristics describing a given painting. The concept vocabulary we use is contains shown below: 

\begin{itemize}
\item \textbf{Style}, caption template: \emph{\{\} style}. Descriptors: 
\begin{itemize}
\item \emph{realism, impressionism, romanticism, expressionism, post impressionism, art nouveau modern, baroque, symbolism, surrealism, neoclassicism, na{\"\i}ve art primitivism, northern renaissance, rococo, cubism, ukiyo e, abstract expressionism, mannerism late renaissance, high renaissance, magic realism, neo impressionism}
\end{itemize}
\item \textbf{Genre}, caption template: \emph{the genre of \{\}}. Descriptors: 
\begin{itemize}
\item \emph{portrait, landscape, genre painting, religious painting, cityscape, sketch and study, illustration, abstract art, figurative, nude painting, design, still life, symbolic painting, marina, mythological painting, flower painting, self portrait, animal painting, photo, history painting, digital art}
\end{itemize}
\item \textbf{Colors}, caption template: \emph{\{\} colors}. Descriptors: 
\begin{itemize}
\item \emph{pale red, pale blue, pale green, pale brown, pale yellow, pale purple, pale gray, black and white, dark red, dark blue, dark green, dark brown, dark yellow, dark purple, dark gray}
\end{itemize}
\item \textbf{Shapes}, caption template: \emph{\{\}}. Descriptors: 
\begin{itemize}
\item \emph{circles, squares, straight lines, rectangles, triangles, curves, sharp angles, curved angles, cubes, spheres, cylinders, diagonal lines, spirals, swirling lines, radial symmetry, grid patterns}
\end{itemize}
\item \textbf{Common Objects}, caption template: \emph{\{\}}. Descriptors: 
\begin{itemize}
\item \emph{male figures, female figures, children, farm animals, pet animals, wild animals, geometric shapes, fruit, vegetables, intsruments, flowers, boats, waves, roads, household items, the moon, the sun, saints, angels, demons}
\end{itemize}
\item \textbf{Backgrounds}, caption template: \emph{\{\} in the background}. Descriptors: 
\begin{itemize}
\item \emph{fields, blue sky, night sky, sunset or sunrise, forest, rolling hills, simple colors, beach, port, river, starry night, clouds, shadows, living room, bedroom, trees, buildings, chapels, heaven, hell, houses, streets}
\end{itemize}
\item \textbf{Color Palette}, caption template: \emph{\{\} color palette}. Descriptors: 
\begin{itemize}
\item \emph{vibrant, muted, monochromatic, complementary, pastel, bright, dull, earthy, bold, subdued, rich, simple, complex, varying, minimal, contrasting}
\end{itemize}
\item \textbf{Medium}, caption template: \emph{the medium of \{\}}. Descriptors: 
\begin{itemize}
\item \emph{oil painting, watercolor, acrylic, ink, pencil, charcoal, etching, screen printing, relief, intaglio, collage, montage, photography, sculpture, ceramics, glass}
\end{itemize}
\item \textbf{Cultural Influence}, caption template: \emph{\{\} influences}. Descriptors: 
\begin{itemize}
\item \emph{Indigenous, European, American, East Asian, Indian, Middle Eastern, Hispanic, Aztec, Contemporary, Greek, Roman, Byzantine, Russian, African, Egyptian, Tahitian, Polynesian, Dutch}
\end{itemize}
\item \textbf{Texture}, caption template: \emph{\{\} texture}. Descriptors: 
\begin{itemize}
\item \emph{rough, smooth, bumpy, glossy, matte, roughened, polished, textured, smoothed, brushstroked, layered, scraped, glazed, streaked, blended, uneven, smudged}
\end{itemize}
\item \textbf{Other Elements}, caption template: \emph{\{\}}. Descriptors: 
\begin{itemize}
\item \emph{stippled brushwork, chiaroscuro lighting, pointillist brushwork, multimedia composition, impasto technique, repetitive, pop culture references, written words, chinese characters, japanese characters}
\end{itemize}
\end{itemize}

Now, we detail the implementation of our modified zero-shot classification. Recall that in zero-shot classification, one computes a text embedding per class, which amounts to the classification head, and computes an image embedding for the test input, so that the prediction is the class who's text embedding has the highest cosine similarity to the test image embedding. In computing the text embeddings, we take each descriptor (e.g. \emph{Dutch}) and place it an aspect-specific caption template (e.g. \emph{Dutch} $\rightarrow$ \emph{Dutch influences}), and then average embedddings over multiple prompts (e.g. ``artwork containing \emph{Dutch influences}'', ``a piece of art with \emph{Dutch influences}'', etc), as done in \cite{radford2021learning}. We modify standard zero-shot classification to allow for the fact that more than one descriptor (or perhaps none) from a given aspect may be present. Namely, instead of assigning the most similar descriptor per-aspect, we assign an atomic tag for any descriptor who's similarity is significantly higher than other descriptors for that aspect. We achieve this via z-score thresholding: per-aspect, we convert similarities to z-scores by subtracting away the mean and dividing by the standard deviation, and then admit atomic tags who's z-score is at least $1.5$.

The template prompts we utilize for embedding each concept caption are as follows:
\begin{itemize}
\item art with {}
\item a painting with {}
\item an image of art with {}
\item artwork containing {}
\item a piece of art with {}
\item artwork that has {}
\item a work of art with {}
\item famous art that has {}
\item a cropped image of art with {}
\end{itemize}

\subsection{From Image Tags to \emph{unique} Artist Tags}
\label{app-sec:tag_comp}

Recall that we define styles not per-image, but over a set of images. Namely, we seek to surface tags that occur frequently. The best way to do so is to simply count the occurrences of each tag, and discard the ones that rarely appear. However, each atomic tag is not particularly unique with respect to artists. We utilized \emph{efficient composition} of atomic tags to arrive at more unique tag signatures, as shown in figure \ref{fig:tag_composition} and detailed in algorithm \ref{alg:cap}. Importantly, we utilize a threshold here to differentiate what a common tag is; we require a tag to appear in at least three works for an artist in order for the tag to count as a frequently used tag by the artist. We note that tag composition can be done efficiently because we have a relatively low number of tags per image: on average, there are $6.2$ atomic tags per image. Moreover, because the number of occurrences for a composed tag is bound belo by the number of occurrences of each atomic tag in the composition, we can ignore all non-frequent atomic tags. Thus, we can iterate over the powerset of common atomic tags per image without it taking exorbitantly long. We include one fail safe, which is that in the rare instance where an image has a very high number of common atomic tags, we truncate the tag list to include only $25$ tags. Over the $91k$ images that we encounter, this happens only once. We highlight that our tag composition takes inspiration from \cite{rezaei2023prime}.

\subsection{Predicting Artistic Styles based on Matched Tags}
\label{app-sec:tag_match_inference}

\looseness=-1
Once we have converted tags per image to tags per artist, we can then utilize these artist tags to perform inference over a set of images. Namely, given a test set of images, we extract common tags (including tag compositions) for the test set and compare them to tags extracted for each artist in our reference corpus. Then, we predict the reference artist who shares the most unique tags with the test set. 

\begin{figure}[t]
    \centering
    \includegraphics[width=\linewidth]{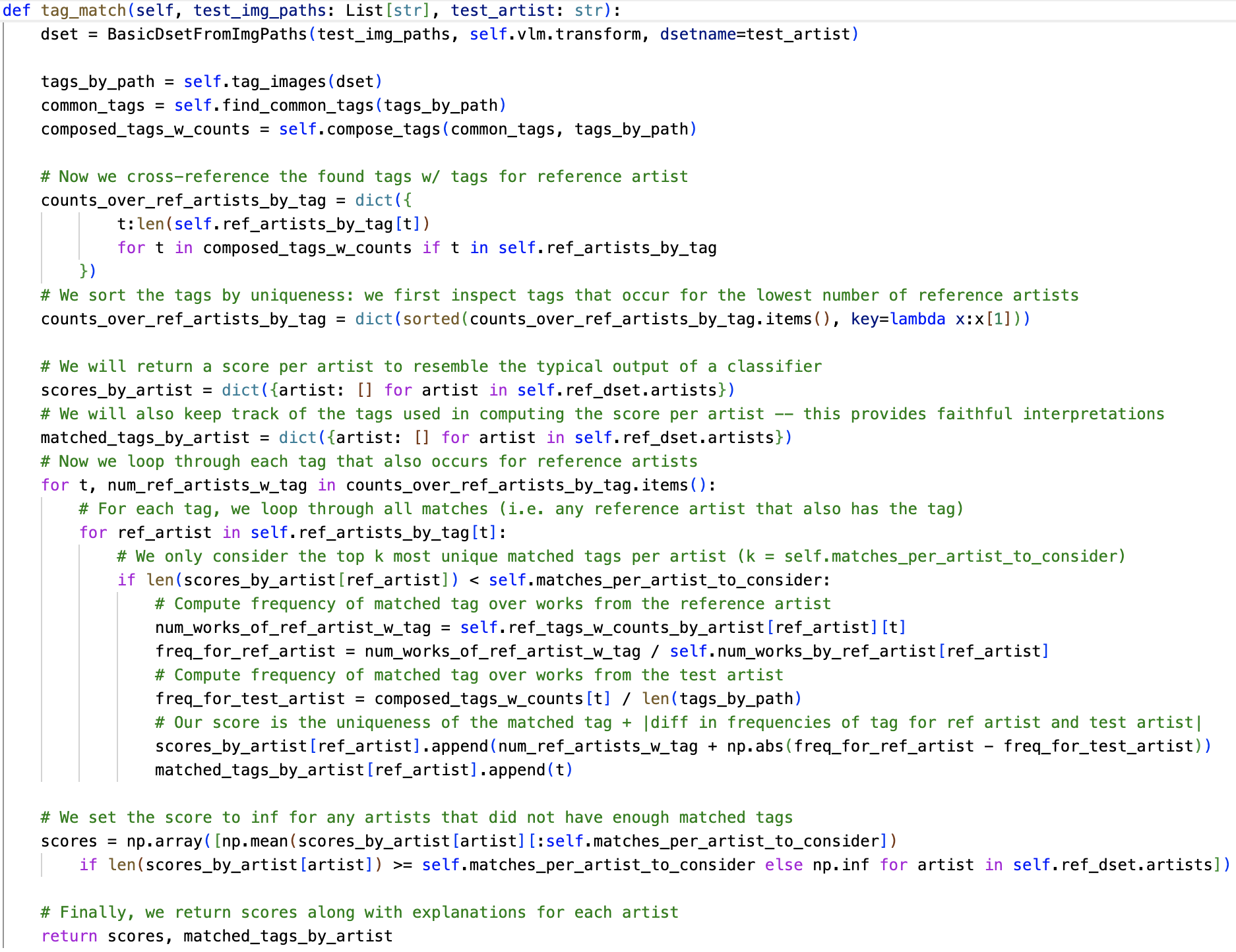}
    \caption{Code for predicting artistic styles via matched tags.}
    \label{fig:tag_match_alg}
\end{figure}

Figure \ref{fig:tag_match_alg} best explains our method, as it shows the documented code. We note that all code will be released upon acceptance. We'll now explain it step by step. First, for each artist and for the test set of images, we find common tags via (i) assigning atomic tags to each image, (ii) finding the commonly occurring atomic tags, (iii) counting compositions of the commonly occurring atomic tags, and (iv) discarding tags (including compositions) that do not occur frequently enough. The code shows this done for the test set of images; we perform this per reference artist when the \texttt{TagMatcher} object (for which \texttt{tag\_match} is function) is initialized; notice fields like \texttt{self.ref\_tags\_w\_counts\_by\_artist}, which contain useful information about the reference artists, computed once and re-used for each inference.

\begin{figure}[t]
    \centering
    \includegraphics[width=\linewidth]{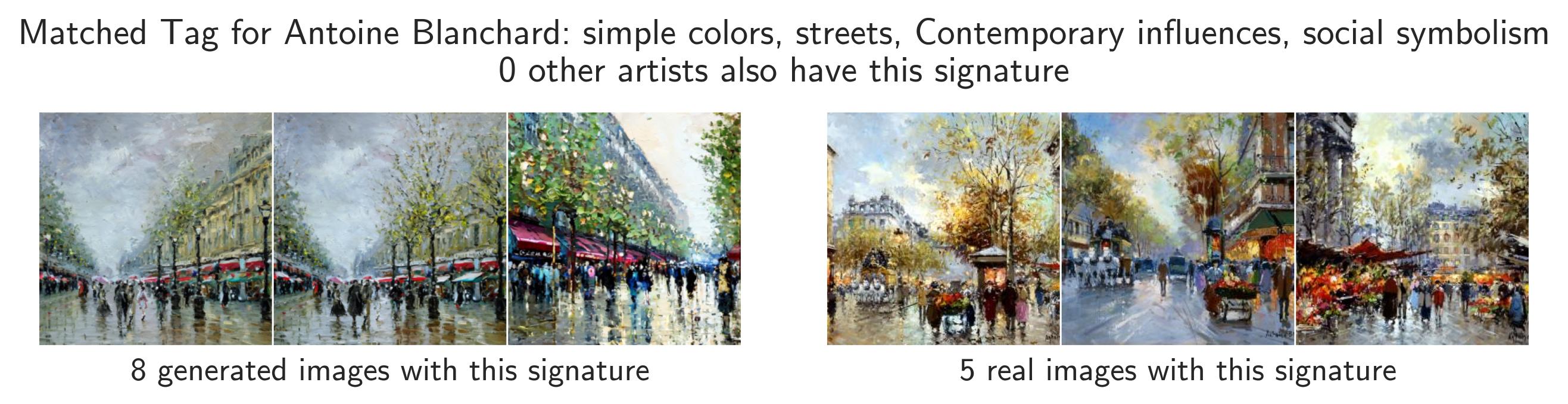}
    \includegraphics[width=\linewidth]{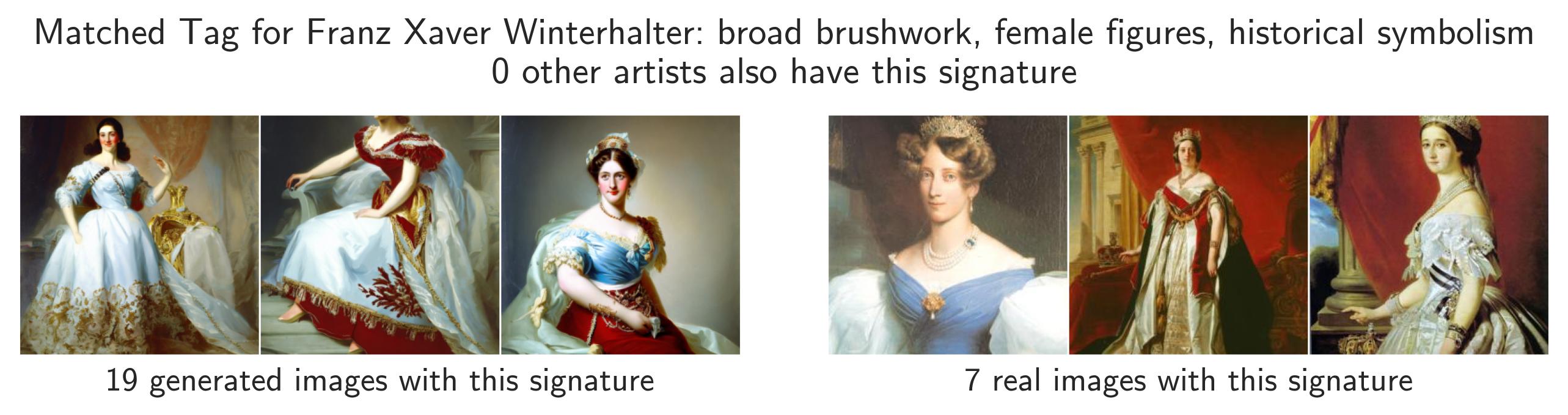}]
    \includegraphics[width=\linewidth]{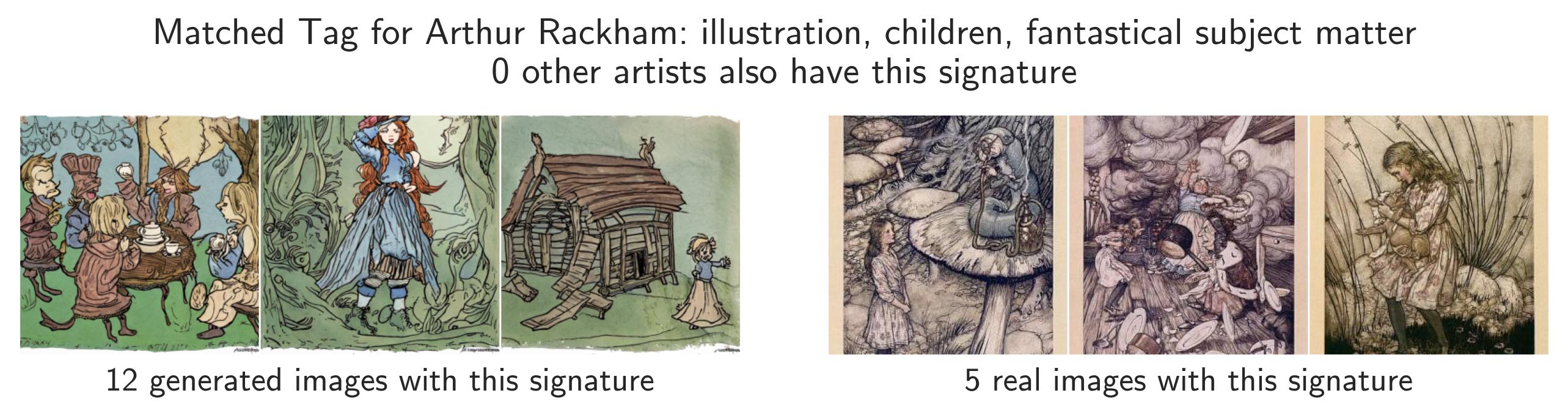}
    \caption{Additional examples of applying TagMatch to generated images.}
    \label{app-fig:tag_match}
\end{figure}

Then, we loop through the set of `matched' tags (i.e. those that occur for both the test set of images and at least one reference artist), starting with the most unique ones. Here, uniqueness refers to the number of reference artists that frequently use a tag. For each tag, we loop through all artists that also use that tag. For the first $k$ (denoted by \texttt{self.matches\_per\_artist\_to\_consider} in the code) matched tags per artist, we add a score to a list of scores for the artist, which ultimately are averaged. The score contains an integer and a decimal component. The integer component is the number of reference artists that share the matched tag. The decimal component is the absolute value of the difference in frequency with which the tag appears, over the reference artist's works and the test set of images; note that this is always less than one. This way, when comparing two matched tags, a lower score is assigned to a more unique one, and one there is a tie in uniqueness, we break the tie based on how similar the frequency of the matched tag is for the test artist and reference artist. 

Finally, we average the list of scores per artist to get a single score per reference artist, analogous to a logit. We assign a score of \texttt{inf} for any artist with less than \texttt{self.matches\_per\_artist\_to\_consider} (which we set to $10$) matched tags. This hyperparameter makes our tag matching less sensitive to individual matched tags, and empirically results in a substantial improvement in top-1 accuracy on held-out art from WikiArt artists (see next section). 

\subsection{Choosing Hyperparameters}
\label{app-sec:hyperparams}

Overall, there are three hyperparameters to our method: the z-score threshold, the tag count threshold, and the number of matches to consider per artist. Here is quick refresher on what they each do:

\begin{itemize}
    \item The z-score threshold determines how much more similar a descriptor needs to be to an image compared to other descriptors for the same aspect in order for the descriptor to be assigned as an atomic tag of the image. The value we use is $1.5$.
    \item The tag count threshold is the minimum number of an artist's works that a tag needs to be present in order for a the tag to be deemed common for the artist. The value we use is $3$. 
    \item The number of matches to consider per artist pertains to how many matched tags are considered when computing the final score per artist in tag match. That is, the final score for an artist is the average of the top-k most unique tags that the artist shares with the test set of images, where $k$ corresponds to this hyperparameter. The value we use is $10$. 
\end{itemize}

Now that the role of each hyperparameter is clear, let's discuss how hyperparameters can be adjusted towards particular ends, along with the potential consequence of each action:

\begin{itemize}
    \item To increase the number of atomic tags, lower the z-score threshold. Risk: atomic tags may be less precise, and the method will take longer to run, as there will atomic tags and composed tags.
    \item To get more tags per artist, lower the tag count threshold. Risk: some tags will become less unique. Other tags will be introduced, and may be very unique, which could skew tag matching. Also, the method may take longer to run, as there will be more tags. 
    \item To make inference less sensitive to a low number of matched tags, increase the number of matches to consider per artist. Risk: when you consider more matches, interpretation is a little more difficult, as you have more reasons for each inference, and it will take longer to view them all. 
\end{itemize}

% TODO: do small sweep of tag match hyperparams

\begin{figure}[t]
    \centering
    \includegraphics[width=\linewidth]{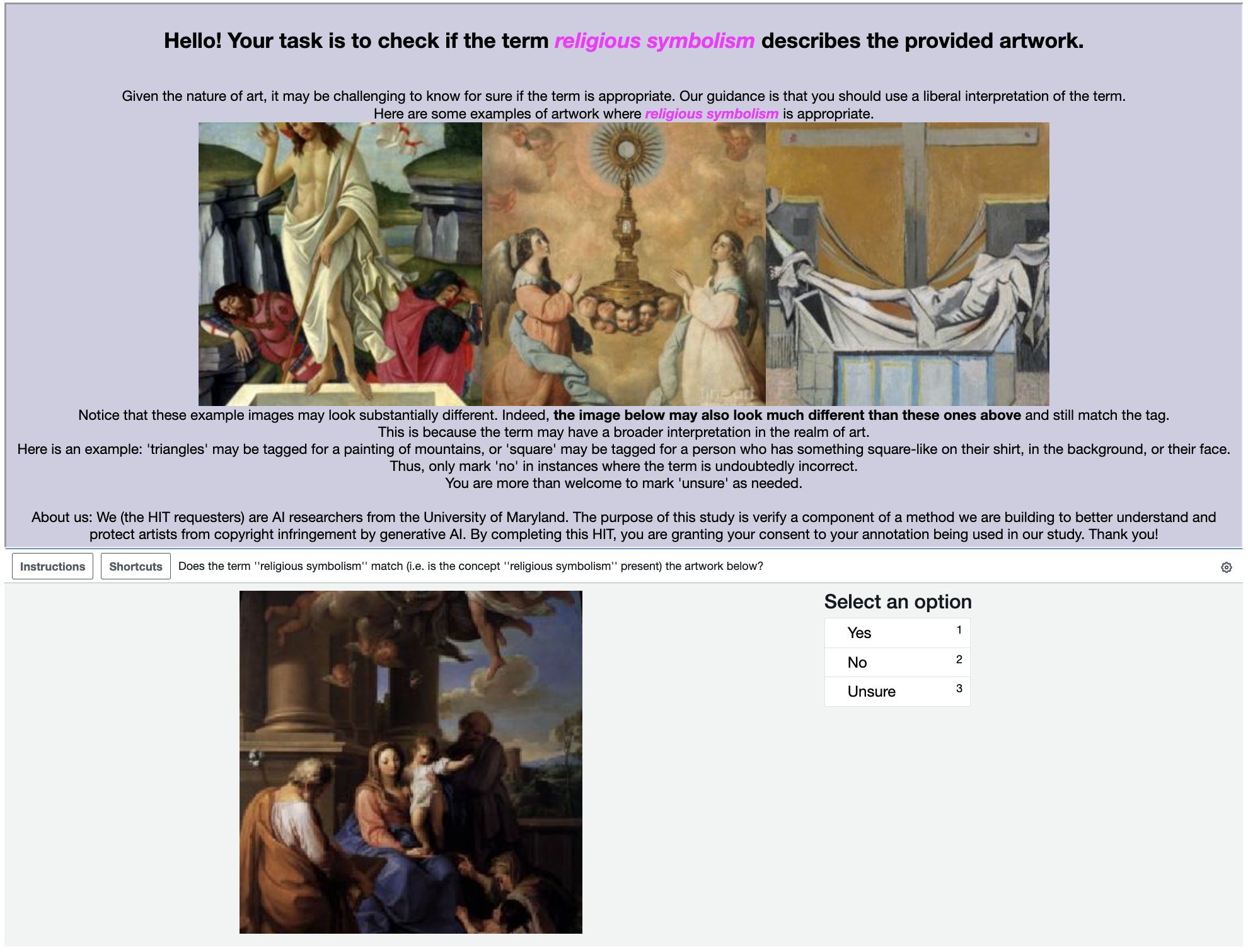}
    \caption{Instructions showed to MTurk workers to validate atomic tags.}
    \label{app-fig:mturk}
\end{figure}

To choose hyperparameters, we selected a small range of reasonable values and swept each hyperparameter individually. While a combined search would likely yield better accuracy numbers, we opt out of hyper-tuning TagMatch for accuracy, as its main objective is to provide and interpretable and attributable complement to DeepMatch. We find the (relatively strong, considering the high number of artists considered) accuracy numbers encouraging, but do not find it a priority, as DeepMatch arguably provides a stronger and easier to understand signal of \emph{if} style copying is happening. TagMatch, on the other hand, tells us \emph{how} and \emph{where} it is happening (if observed with DeepMatch).

% We also include a small search of each hyperparameter individually, visualized in Figure \ref{fig:hyperparam_sweeps}. We note that the main goal of TagMatch is not to be super accurate, but to complement DeepMatch with interpretations (via matched tag signatures) and attributions (via works from the test set and from the reference artist that present the matched tags). 

\subsection{Efficiency of TagMatch: Runs in $\sim 1$ minute}
\label{app-sec:time_for_tagmatch}

TagMatch is surprisingly fast. The longest step by far is computing CLIP embeddings for the reference artworks. This takes us about 5 minutes using one rtx2080 GPU with four CPU cores to embed the $73k$ training split images using a CLIP ViT-B$\backslash$16 model. Importantly, this step is done only once, and in practice, is done offline. The other steps and approximate time needed for each are as follows: embedding concepts (5 seconds), extracting common atomic tags and composing them (45 seconds), reorganizing tags and removing non-common tags (3 seconds). Then, inference for a test set of $100-200$ works takes about 10 to 15 seconds. Again, we will release all code upon acceptance, as we truly hope our tool can be of use to artists who are concerned by generative models potential infringing upon their unique styles. 

\subsection{Validation}
\label{app-sec:tagmatch_val}

Because tag match has multiple steps, we perform multiple validations. First, for image tagging, we utilize an MTurk study. We collect $3000$ separate human judgements on instances of assigned atomic tags. Namely, we show $1000$ randomly selected (tag, image) pairs to three annotators each. Figure \ref{app-fig:mturk} shows an example of the form presented to MTurk workers. MTurkers provide consent and are awarded $\$0.15$ per task, resulting in an estimated hourly pay of $\$12-\$18$. For each task, they answer `yes', `no', or `unsure' to the question `does the term \{atomic tag\} match the artwork below?' They are also shown example artworks for each term which were manually verified to be correct. Response rates were as follows: $69.89\%$ yes, $8.99\%$ unsure, $21.12\%$ no. In investigating inter-annotator agreement, we find that at least 2 annotators agree $92.1\%$ of the time, but all 3 agree only $51.52\%$ of the time. This reflects the subjectivity associated with assigning artistic tags, and partially motivates the need for a deterministic automated alternative, in order to objectively tag images at scale. All three annotators said no only $5.16\%$ of the time, and at least two said no $17.11\%$ of the time, suggesting that our zero-shot tagging mechanism achieves reasonable precision.

To validate the value of tag composition, we refer to figure \ref{fig:tag_composition}, which shows how tags become more unique as they get longer (i.e. consist of more atomic tags). Moreover, our time analyses show that the added benefit of composing tags to find unique tag signatures does not come at the cost of the efficiency of our method. Finally, the non-trivial top-1 matching accuracy and strong top-5 matching accuracy shows that the extracted tag signatures do indeed capture some unique properties of artistic style. Figure \ref{app-fig:tag_match} reflects a few more examples of successful inference, interpretation, and attribution for the task of detecting style copying by generative models.

\section{A Sim2Real Gap in Tag Distributions}

An added advantage of ascribing tags to images is that we can better compare image distributions from an interpretable basis (the tags). We briefly explore this direction now. 

\begin{table}[t]
    \centering
    \begin{tabular}{llrrr}
    \toprule
    {} & {} &   Top 1 &   Top 5 &  Top 10 \\
    \midrule
    % Generated Art & Model & & & \\
    % & CompVis Stable Diffusion v1.4    & 10.10\% & 35.49\% & 49.74\% \\
    % & Stability AI Stable Diffusion v2 & 12.95\% & 37.82\% & 52.59\% \\
    % & PromptHero Openjourney           &  6.99\% & 31.87\% & 45.08\% \\
    Generated Art & CompVis Stable Diffusion v1.4    & $10.10$ & $35.49$ & $49.74$ \\
& Stability AI Stable Diffusion v2 & $12.95$ & $37.82$ & $52.59$ \\
& PromptHero Openjourney           &  $6.99$ & $31.87$ & $45.08$ \\
    
    & Average &                          $10.02$ & $35.06$ & $49.14$ \\
    \midrule
    Real Art (held out) & & $60.05$ & $80.68$ & $86.86$ \\
    \bottomrule
    \end{tabular}
    \caption{Match rates using TagMatch for three generative models, as well as on real held out art.}
    \label{tab:tag_match_full_results}
\end{table}

First, we provide complete results from applying TagMatch to generated images from each of the three text-to-image models in our study, presented in table \ref{tab:tag_match_full_results}. Consistent with our DeepMatch results, we observe substantially lower matching accuracy for generated images than for real held-out artwork. While the primary takeaway is that for many artists, generative models struggle to replicate their styles, we can also hypothesize that generative models may output images that follow a different distribution than the distribution of real artworks.

Motivated by this hypothesis, we now compare the distribution of real to 
\begin{wrapfigure}{r}{0.5\textwidth}
    \centering
    \includegraphics[width=\linewidth]{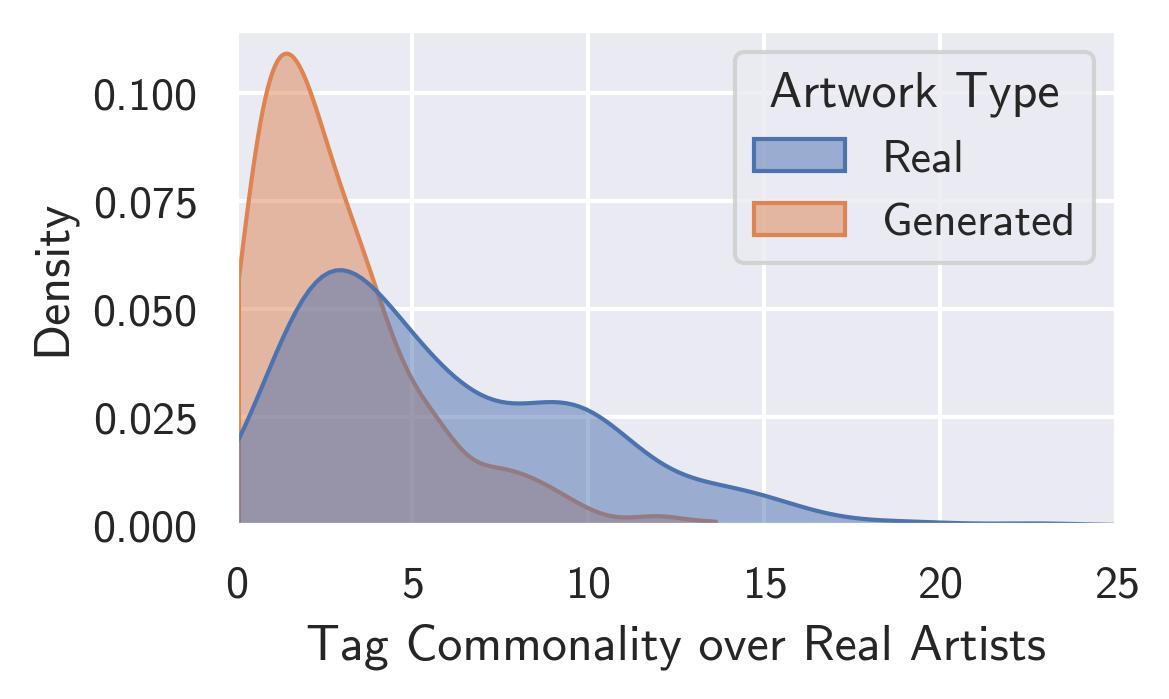}
    \caption{The tags for generated images are less common compared to tags in real art. }
    \label{fig:tag_commonality}
\end{wrapfigure}
generated artworks from the perspective of tags. Because we consider composed tags, the total space of tags is vast and hard to reason over. However, we can look at properties of each tags. Namely, we can inspect the uniqueness of tags. That is, for each tag present in generated images, we inspect the number of reference artists that also present that tag; we do the same for real art as well (subtracting one so to not double count the artist for which a given a tag is being considered). Figure \ref{fig:tag_commonality} shows a kernel density estimation plot of the distributions of tag commonality, where a tag commonality of $5$ means that for each tag assigned to a set of images (either from a real artist or from a generative model emulating an artist), $5$ other artists also commonly use that tag. We see tags tend to be rather unique (due to our tag composition), and notably, tags for generated images are more unique.

\section{Patch Match: Generating Additional Visual Evidence of Copying}

% Atoosa's section. Should include:

% \begin{itemize}
%     \item qualitative discussion of clip and gram matrix based patch match; pros and cons of each; ideally a small figure here too 
%     \item how long the method takes, and explain since can be more costly, we only use it to search for direct copied patches for the artist of interest
%     \item A *good* example for Van Gogh (or anyone, but ideally van gogh -- we should really be able to match the spirals that these models keep generating).
%     \item A discussion of limitations, referring back to our earlier discussion of how image-wise similarities doesn't capture the whole picture. 
% \end{itemize}

Detecting artistic style copying in a given art requires analyzing local stylistic elements that manifest across an artist's body of work. To address this, we employ a patch-based approach that compares small image regions between a given art and original artworks, enabling a fine-grained analysis of stylistic and semantic (e.g. objects) similarities at a local level. We consider three patch matching methods: CLIP-based, DINO-based, and Gram matrix-based. \\

\textbf{Gram Matrix-based Patch Matching \cite{gatys2016image}}: The Gram matrix is a measure of style similarity introduced in the context of neural style transfer. It captures the correlations between the activations of different feature maps in a convolutional neural network, representing the style of an image. For patch matching, the Gram matrices of patches from the given art and original arts can be computed and compared using a suitable distance metric (e.g., Frobenius norm). The Gram matrix is specifically designed to capture stylistic elements, making it well-suited for detecting style copying.

\textbf{CLIP-based Patch Matching \cite{radford2021learning}}: CLIP (Contrastive Language-Image Pre-training) is a powerful model that can effectively capture the semantic similarity between text and images. In the context of patch matching, CLIP embeddings can be used to measure the similarity between a patch from a given art and patches from original artworks. The patches can be encoded using the CLIP image encoder, and the cosine similarity between their embeddings can be computed to find the closest matches. CLIP may not be as sensitive to low-level stylistic elements, such as brushstrokes, textures, and color palettes, however it focuses more on higher-level semantic concepts, which can be useful to find if the given art pictured the same objects as the selected original patch.

\textbf{DINO-based Patch Matching \cite{caron2021emerging}}: DINO is a self-supervised vision transformer that learns robust visual representations by solving a self-distillation task. DINO embeddings can be used for patch matching by computing the cosine similarity between the embeddings of patches from the given art and original artworks. We use DINO to capture higher semantical similarities, and check whether the given art pictured similar subjects of interest and high-level visual features as selected original artworks.

\subsection{Experimental setting}
For our experiments, we aim to identify the most similar artwork from a pool of $10,000$ original artworks in the WikiArt dataset given a reference image. The reference image is first resized to a resolution of $512*512$ pixels and normalized. From this normalized image, we select a patch size of $128*128$ pixels. This process is repeated for all original artworks in the dataset, resulting in a total of $40,000$ patches from original artworks for comparison with the reference patch. We then use three methods, namely Gram matrix, CLIP, and DINO, to find the most similar patches.

\begin{figure}[t]
    \centering
    \includegraphics[width=0.7\linewidth]{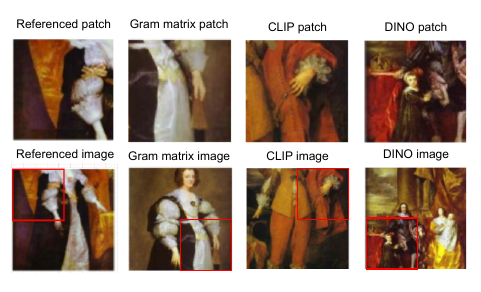}
    \caption{The most similar patches to a referenced patch in an image using Gram-matrix, CLIP, and DINO.}
    \label{fig:patch_match1}
\end{figure}

Figure \ref{fig:patch_match1} showcases the patches that are deemed most similar to the image being referenced. These matches are determined using Gram-matrix, CLIP, and DINO methods.

\begin{figure}[t]
    \centering
    \includegraphics[width=\linewidth]{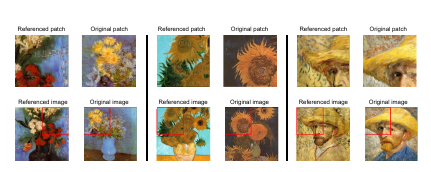}
    \caption{Comparison of patches using the Gram-matrix method, highlighting the closest matches to three selected artworks by Van Gogh. The selected original arts, all from Van Gogh, closely resemble the style of the referenced paintings.}
    \label{fig:patch_match2}
\end{figure}

We then select an artist and find patches from our original image dataset that closely match this artist's style. In Figure \ref{fig:patch_match2}, we utilize the Gram-matrix method to identify the most similar patches to three chosen artworks by Van Gogh. Our dataset includes all paintings by Van Gogh as well as works by nine other artists. Gram-matrix selects original artworks that closely resemble the style of the reference image, all of which are from Van Gogh. Essentially, this means that Gram-matrix predominantly selects Van Gogh's artworks because they are the most stylistically similar to the referenced paintings compared to the works of the other nine artists.

\subsection{Discussion and limitations}
Patch matching methods like Gram-matrix, CLIP, and DINO are effective in detecting similarities between artworks by examining their local stylistic and semantic elements. Gram-matrix focuses on capturing stylistic correlations, CLIP evaluates semantic similarity, and DINO concentrates on higher-level features. However, these methods have limitations. They primarily focus on local aspects of artworks and may overlook broader artistic characteristics such as texture, composition, and brushwork that are crucial to detect copyright infringements. Moreover, the process of finding the most similar patches for each given art takes approximately fifteen minutes when considering $10,000$ original artworks, and if we opt to include more original artworks, the duration of the process would inevitably increase. Therefore, patch-matching methods are computationally expensive, which restricts their practical application. Despite these limitations, patch matching is valuable for identifying instances of direct copying in artworks and they aid in the detection of plagiarized content.

\section{Details on WikiArt Scraping}

WikiArt is a free project intended to collect art from various institutions, like museums and universities, to make them readily accessible to a broader audience. We design a scraper to collect a corpus of reference artists, with which we can define a test artist's style in contrast to the other artists, and to provide a testbed to empirically study copying behavior of generative models. Some important landing pages to perform scraping are (i) the works by artist page (\url{https://www.wikiart.org/en/Alphabet/j/text-list}; url shows all artists starting with the letter `j', and we loop through all letters), (ii) the page containing information on allowed usage (\url{https://www.wikiart.org/en/terms-of-use}), (iii) an example artist landing page (\url{https://www.wikiart.org/en/vincent-van-gogh}), and (iv) an example painting landing page (\url{https://www.wikiart.org/en/vincent-van-gogh/the-starry-night-1889}). As you can see, many pages have standard formats, making scraping particularly feasible. We will provide our scraping code, along with all other code, to facilitate easy updating of our dataset as time goes by. 

We obtain artworks only from artists with at least 100 works on WikiArt, so to focus on somewhat famous artists who are arguably more likely to be copied. For every work, we also scrape the licensing information, and annotation for styles, genres, and title. In total, our dataset has 90,960 artworks over 372 artists. There are 81 styles with at least 100 works, with the most popular styles being \emph{realism, impressionism, romanticism,} and \emph{expressionism}. There were 37 genres with at least 100 works, with the most popular being \emph{portrait, landscape, religious painting, sketch and study}, and \emph{cityscape}. We note that we only include images who's license is either public domain or fair use, with the vast majority of works being public domain. Nonetheless, we strongly advise against using this dataset for commercial purposes, and especially for the purpose of copying artists.

\end{document}